\documentclass[review]{elsarticle}

\makeatletter
\def\ps@pprintTitle{%
 \let\@oddhead\@empty
 \let\@evenhead\@empty
    \def\@oddfoot{\footnotesize\itshape
         {under review} \hfill\today}%
 \let\@evenfoot\@oddfoot}
\makeatother

\usepackage{amsmath}
\usepackage{amsfonts}
\usepackage{amssymb}
\usepackage{amsthm}
\usepackage{bm}
\usepackage{indentfirst}
\usepackage{graphicx}
\usepackage{subfigure}
\usepackage{array}
\usepackage{fullpage}

\usepackage{amsmath,amsfonts,bm}

\usepackage{amsmath,amssymb,amsfonts,mathrsfs}
\usepackage{hyperref}
\usepackage{url}
\usepackage{bm}
\usepackage{amsthm}
\usepackage{algorithm}
\usepackage{algorithmic}
\biboptions{sort&compress}
\usepackage{graphicx}

\def\eqref#1{equation~\ref{#1}}

\def\1{\bm{1}}

\def\vb{{\bm{b}}}

\DeclareMathAlphabet{\mathsfit}{\encodingdefault}{\sfdefault}{m}{sl}
\SetMathAlphabet{\mathsfit}{bold}{\encodingdefault}{\sfdefault}{bx}{n}

    \newtheoremstyle{TheoremNum}
        {\topsep}{\topsep}              
        {\itshape}                      
        {}                              
        {\bfseries}                     
        {.}                             
        { }                             
        {\thmname{#1}\thmnote{ \bfseries #3}}
    \theoremstyle{TheoremNum}
    \newtheorem{thmn}{Theorem}

\def\omg{{\Omega}}
\newcommand{\mcU}{\mathcal{U}}
\newcommand{\mcJ}{\mathcal{J}}

\newcommand{\mcS}{\mathcal{S}}
\newcommand{\mcK}{\mathcal{K}}
\newcommand{\mcB}{\mathcal{B}}
\newcommand{\mcR}{\mathcal{R}}

\newcommand{\mcD}{\mathcal{D}}
\newcommand{\mcG}{\mathcal{G}}
\newcommand{\mcH}{\mathcal{H}}

\newcommand{\mcF}{\mathcal{F}}

\newcommand{\mcL}{\mathcal{L}}

\newcommand{\mcA}{\mathcal{A}}
\newcommand{\mcT}{\mathcal{T}}

\newcommand{\mcZ}{\mathcal{Z}}

\def \bb{\mathbf{b}}

\def \Fb{\mathbf{F}}
\def \Cb{\mathbf{C}}
\def \Hb{\mathbf{H}}
\def \Gb{\mathbf{G}}

\def \gb{\mathbf{g}}
\def \ub{\mathbf{u}}
\def \Ub{\mathbf{U}}

\def \vb{\mathbf{v}}

\def \xb{\mathbf{x}}
\def \Xb{\mathbf{X}}

\def \hb{\mathbf{h}}
\def \pb{\mathbf{p}}
\def \qb{\mathbf{q}}

\def \cb{\mathbf{c}}

\newcommand{\real}{\mathbb{R}}
\newcommand{\complex}{\mathbb{C}}

\newcommand{\vertii}[1]{{\left\vert\left\vert #1
    \right\vert\right\vert}}

\usepackage{microtype}
\usepackage{graphicx}
\usepackage{subfigure}
\usepackage{booktabs} 

\usepackage{hyperref}

\theoremstyle{plain}
\newtheorem{theorem}{Theorem}[section]

\newtheorem{lemma}[theorem]{Lemma}

\theoremstyle{definition}

\newtheorem{assumption}[theorem]{Assumption}
\theoremstyle{remark}

\usepackage[textsize=tiny]{todonotes}

\begin{document}

\begin{frontmatter}

\title{MetaNO: How to Transfer Your Knowledge on Learning Hidden Physics}

\address[yy]{Department of Mathematics, Lehigh University, Bethlehem, PA, USA}
\address[tg]{IBM Research, Yorktown Heights, NY, USA}
\address[my]{Pattern Recognition Center, WeChat AI, Tencent Inc, China}
\address[cl]{School of Aerospace and Mechanical Engineering, The University of Oklahoma, Norman, OK, USA}

\author[yy]{Lu Zhang}
\author[yy]{Huaiqian You}
\author[tg]{Tian Gao}
\author[my]{Mo Yu}
\author[cl]{Chung-Hao Lee}
\author[yy]{Yue Yu\corref{cor1}}\ead{yuy214@lehigh.edu}
\cortext[cor1]{Corresponding author}

\begin{abstract}
Gradient-based meta-learning methods have primarily been applied to  classical machine learning tasks such as image classification.
Recently, PDE-solving deep learning methods, such as neural operators, are starting to make an important impact on learning and predicting the response of a complex physical system directly from observational data. 
Since the data acquisition in this context is commonly challenging and costly, the call of utilization and transfer of existing knowledge to new and unseen physical systems is even more acute. 
Herein, we propose a novel meta-learning approach for neural operators, which can be seen as transferring the knowledge of solution operators between governing (unknown) PDEs with varying parameter fields. Our approach is a provably universal solution operator for multiple PDE solving tasks, with a key theoretical observation that underlying parameter fields can be captured in the first layer of neural operator models, in contrast to typical final-layer transfer in existing meta-learning methods. As applications, we demonstrate the efficacy of our proposed approach on PDE-based datasets and a real-world material modeling problem, illustrating that our method can handle complex and nonlinear physical response learning tasks while greatly improving the sampling efficiency in unseen tasks. 
\end{abstract}


\end{frontmatter}







\section{Introduction}

Few-shot learning is an important problem in machine learning, where new tasks are learned with a very limited number of labelled datapoints~\citep{wang2020generalizing}. In recent years, significant progress has been made on few-shot learning using meta-learning approaches~\citep{koch2015siamese,vinyals2016matching,snell2017prototypical,finn2017model,santoro2016meta,antoniou2018train,ravi2016optimization,nichol2018reptile,raghu2019rapid,tripuraneni2021provable,collins2022maml}. Broadly speaking, given a family of tasks, some of which are used for training and others for testing, meta-learning approaches aim to learn a shared multi-task representation that can generalize across the different training tasks, and result in fast adaptation to new and unseen testing tasks. Meta-learning learning algorithms have been successfully applied to conventional machine learning problems such as image classification, function regression, and reinforcement learning, but studies on few-shot learning approaches for complex physical system modeling problems have been limited. 
The call of developing a few-shot learning approach for complex physical system modeling problems is just as acute, while the typical understanding of how multi-task learning should be applied on this scenario is still nascent.

As a motivating example, we consider the scenario of new material discovery in the lab environment, where the material model is built based on experimental measurements of its responses subject to different loadings. Since the physical properties (such as the mechanical and structural parameters)  in different material specimens  vary, the model learnt from experimental measurements on one specimen would have large generalization errors on other specimens. As a result, the data-driven model has to be trained repeatedly with a large number of material specimens, which makes the learning process inefficient. Furthermore, experimental measurement acquisition of these specimens is often challenging and expensive. In some problems, a large amount of measurements are not even feasible. For example, in the design and testing of biosynthetic tissues, performing repeated loading would potentially induce the cross-linking and permanent set phenomenon, which notoriously alter the tissue durability~\citep{zhang2017modeling}.  As a result, it is  critical to learn the physical response model of a new specimen with sample size as small as possible. Furthermore, since many characterization methods to obtain underlying material mechanistic and structural properties would require the use of destructive methods~\citep{misfeld2007heart,rieppo2008practical}, in practice many physical properties are not measured and can only be treated as hidden and unknown variables. Hence, we likely only have limited access to the measurements on the complex system responses caused by the change of these physical properties. 

Supervised operator learning methods are typically used to address this class of problems. They take  a number of observations  on the loading field as input, and try to predict the corresponding physical system response field as output, corresponding to one underlying PDE (as one task). Herein, we consider the meta-learning of multiple complex physical systems (as tasks), such that all these tasks are governed by a common PDE with different (hidden) physical property or parameter fields.   Formally, assume that we have a distribution $p(\mcT)$ over tasks, each task $\mcT^\eta\sim p(\mcT)$ corresponds to a hidden physical property field $\bb^\eta(\xb)\in\mcB(\real^{d_b})$ that contains the task-specific mechanistic and structural information in our material modeling example. On task $\mcT^\eta$, we have a number of observations on the loading field $\gb_i^\eta(\xb)\in\mcA(\real^{d_g})$ and the corresponding physical system response field $\ub_i^\eta(\xb)\in\mcU(\real^{d_u})$ according to a hidden parameter field $\bb^\eta(\xb)$. Here, $i$ is the sample index, $\mcB$, $\mcA$ and $\mcU$ are Banach spaces of function taking values in $\real^{d_b}$, $\real^{d_g}$ and $\real^{d_u}$, respectively. 
For task $\mcT^\eta$, our modeling goal is to learn the solution operator $\mcG^\eta:\mcA\rightarrow\mcU$, such that the learnt model can predict the corresponding physical response field $\ub(\xb)$ 
for any loading field $\gb(\xb)$. Without transfer learning, one needs to learn a surrogate solution operator for each task only based on the data pairs on this task, and repeat the training for every task. The learning procedure would require a relatively large amount of observation pairs and training time for each task. Therefore, this physical-based modeling scenario raises a key question: \textit{ {Given data from a number of parametric PDE solving (training) tasks with different unknown parameters},  how can one efficiently learn an accurate surrogate solution operator for a test task with new and unknown parameters, with few data on this task\footnote{In some meta-learning literature, e.g., \citep{xu2020metafun}, these small sets of labelled data pairs on a new task (or any task) is called the context, and the learnt model will be evaluated on an additional set of unlabelled data pairs, i.e., the target.}?}

To address this question, we introduce MetaNO, {a novel meta-learning approach for transferring knowledge between neural operators}, which can be seen as transferring the knowledge of solution operators between governing (potentially unknown) PDEs with varying hidden parameter fields. Our \textbf{main contributions} are:
\begin{itemize}
    \item MetaNO is the first neural-operator-based {meta-learning approach for multiple tasks}, which not only preserves the generalizability to different resolutions and input functions from the integral neural operator architecture, but also improves sampling efficiency on new tasks -- for comparable accuracy, MetaNO saves the number of measurements required by $\sim$90\%.
    \item With rigorous operator approximation analysis, we made the key observation that the hidden parameter field can be captured by adapting the first layer of the neural operator model. 
    Therefore, our MetaNO is \textit{substantially different} from existed popular meta-learning approaches \cite{finn2017model,raghu2019rapid}, since the later typically rely on the adaptation of their last layers \cite{collins2022maml}. By construction, MetaNO serves as a provably universal solution operator for multiple PDE solving tasks.
    \item On synthetic, benchmark, and real-world biological tissue datasets, the proposed method consistently outperforms {existing non-meta transfer-learning baselines} and  other gradient-based meta-learning methods. 
\end{itemize}

\begin{figure*}[t]
	\centering
\includegraphics[width=1.\columnwidth]{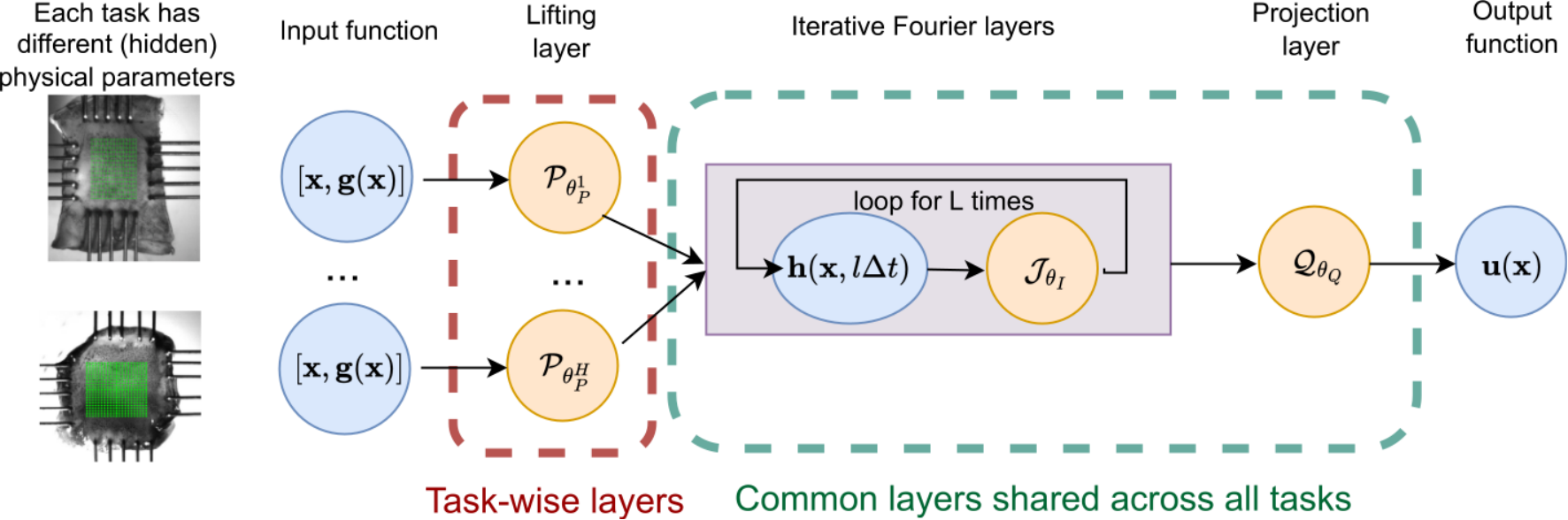}
	\caption{The architecture of MetaNO based on an integral neural operator model.
 }
\label{fig:domain}
\end{figure*}

\section{Background and Related Work}


\subsection{Hidden Physics Learning with Neural Networks}

For many decades, physics-based PDEs have been commonly employed for predicting and monitoring complex system responses. Then traditional numerical methods were developed to solve these PDEs and provide predictions for desired system responses. However, three fundamental challenges usually present. 
First, the choice of governing PDE laws is often determined \textit{a priori} and free parameters are often tuned to obtain agreement with experimental data, which makes the rigorous calibration and validation process challenging. Second, traditional numerical methods are solved for specific boundary and initial conditions, as well as loading or source terms. Therefore, they are not generalizable for other operating conditions and hence not effective for real-time prediction. Third, complex PDE systems such as turbulence flows and heterogeneous materials modeling problems usually require a very fine discretization, and are therefore  very time-consuming for traditional solvers.

To provide an efficient surrogate model for physical responses, machine learning methods may hold the key. Recently, there has been significant progress in the development of deep neural networks (NNs) for learning the hidden physics of a complex system \citep{ghaboussi1998autoprogressive,ghaboussi1991knowledge,carleo2019machine,karniadakis2021physics,zhang2018deep,cai2022physics,pfau2020ab,he2021manifold,besnard2006finite}. Among these methods, the neural operators show particular promises in resolving the above challenges, 
which aim to learn mappings between inputs of a dynamical system and its state, so that the network can serve as a surrogate for a solution operator \citep{li2020neural,li2020multipole,li2020fourier,you2022nonlocal,Ong2022,gupta2021multiwaveletbased,lu2019deeponet,lu2021learning,goswami2022physics}. 

Comparing with classical NNs,  most notable advantages of neural operators are resolution independence and generalizability to different input instances.
Moreover, comparing with the classical PDE modeling approaches, neural operators require only data with no knowledge of the underlying PDE. All these advantages make neural operators promising tools to PDE learning tasks. Examples include modeling the unknown physics law of real-world problems \citep{yin2022simulating,yin2022interfacing} and providing efficient solution operator for PDEs \citep{li2020neural,li2020multipole,li2020fourier,lu2021one,lu2021comprehensive}. 
On the other hand, data in scientific applications are often scarce and incomplete. 
Utilization of other relevant data sources could alleviate such a problem, yet
no existing work have addressed the transferability of neural operators. Through the meta-learning techniques, our work fulfills the demand of such a transfer setting, with the same type of PDE system but different (hidden) physical properties.

\subsection{Base Model: Integral Neural Operators}

We briefly introduce the integral neural operator model, which will be utilized as the base model of this work. The integral neural operators, first proposed in \citep{li2020neural} and further developed in \citep{li2020multipole,li2020fourier,you2022nonlocal,you2022learning} comprises of three building blocks. First, the input function, $\gb(\xb)\in\mcA$, is lifted to a higher dimensional representation via $\hb(\xb,0)=\mathcal{P}[\gb](\xb):=P(\xb)[\xb,\gb(\xb)]^T+\pb(\xb)$. $P(\xb)\in\real^{(s+d_g)\times d_h}$ and $\pb(\xb)\in\real^{d_h}$ define an affine pointwise mapping, which are often taken as constant parameters, i.e., $P(\xb)\equiv P$ and $\pb(\xb)\equiv \pb$. Then, the feature vector function $\hb(\xb,0)$ goes through an iterative layer block where the layer update is defined via the action of the sum of a local linear operator, a nonlocal integral kernel operator, and a bias function:  $\hb(\cdot,l+1)=\mathcal{J}_{l+1}[\hb(\cdot,l)]$. Here, $\hb(\cdot,l)$, {$l\in \{0,\cdots,L\}$}, is a sequence of functions representing  values of the network at each hidden layer, taking values in $\real^{d_h}$. $\mathcal{J}_1,\cdots,\mathcal{J}_{L}$ are  nonlinear operator layers. 
In this work, we employ the implicit Fourier neural operator (IFNO) as the base model
\footnote{We also point out that the proposed multi-task strategy is generic and hence also applicable to other neural operators \citep{lu2019deeponet,li2020neural,li2020multipole,li2020fourier,you2022nonlocal}. } and take the iterative layers as $\mathcal{J}_1=\cdots=\mathcal{J}_{L}=\mathcal{J}$, where
\begin{align}
&\hb(\xb,l+1)=\mcJ[\hb(\xb,l)]:=\hb(\xb,l)+ \dfrac{1}{L}\sigma(W\hb(\xb,l)+\mathcal{F}^{-1}[\mathcal{F}[\kappa(\cdot;\vb)]\cdot \mathcal{F}[\hb(\cdot,l)]](\xb)+ \cb(\xb)).\label{eq:IFNO}
\end{align}
$\mathcal{F}$ and $\mathcal{F}^{-1}$ denote the Fourier transform and its inverse, respectively. 
$\cb\in\real^{d_h}$ defines a constant bias, $W\in\real^{d_h\times d_h}$ is the weight matrix, and $\mathcal{F}[\kappa(\cdot;\vb)]:=R$ is a circulant matrix that depends on the convolution kernel $\kappa$. $\sigma$ is an activation function, which is often taken to be the popular rectified linear unit (ReLU) function. Finally, the output $\ub(\cdot)\in\mcU$ is obtained through a projection layer, by mapping the last hidden layer representation $\hb(\cdot,L)$ onto $\mcU$ as:
$\ub(\xb)=\mathcal{Q}[\hb(\cdot,L)](\xb):=Q_2(\xb)\sigma(Q_1\hb(\xb,L)+\qb_1(\xb))+\qb_2(\xb)$. $Q_1(\xb)\in\real^{d_{Q}\times d_h}$, $Q_2(\xb)\in\real^{d_{u}\times d_Q}$, $\qb_1(\xb)\in\real^{d_Q}$ and $\qb_2(\xb)\in\real^{d_u}$ are  appropriately sized matrices and vectors that are part of the parameter set that we aim to learn, 
which are often taken as constant parameters and will be denoted as $Q_1$, $Q_2$, $\qb_1$ and $\qb_2$, respectively. In the following, we denote the set of trainable parameters in the lifting layer as $\theta_P$, the set from the iterative layer block as $\theta_I$, and the set in the projection layer as $\theta_Q$.

The neural operator can be employed to learn an approximation for the solution operator, $\mcG$. Given $\mcD:=\{(\gb_i,\ub_i)\}_{i=1}^N$, a labelled (context) set of observations, where the input $\{\gb_i\}\subset\mcA$ is a set of independent and identically distributed (i.i.d.) random fields from a known probability distribution $\mu$ on $\mcA$, and $\ub_i(\xb)\in\mcU$ is the observed but possibly noisy corresponding solution. Let $\omg\subset\real^s$ be the domain of interest, we assume that all observations can be modeled with a parametric PDE form:
\begin{equation}\label{eqn:pde_single}
\begin{aligned} 
\mcK_{\bb(\xb)} [\ub_i] (\xb)=\gb_i (\xb),\quad&\xb\in \omg.
\end{aligned}
\end{equation}
$\mcK_\bb$ is the operator representing the possibly unknown governing law, e.g., balance laws. Then, the system response can be learnt by constructing a surrogate solution operator of \eqref{eqn:pde_single}: $\tilde{\mcG}[\gb;\theta](\xb):=\mathcal{Q}_{\theta_Q}\circ(\mathcal{J}_{\theta_I})^L\circ\mathcal{P}_{\theta_P}[\gb](\xb)\approx \ub(\xb)$, where  parameter set $\theta=[\theta_P,\theta_I,\theta_Q]$ is obtained by solving the optimization problem:
\begin{equation}\label{eqn:opt}
\min_{\theta\in\Theta}\mcL_{\mcD}(\theta):=
\min_{\theta\in\Theta}\sum_{i=1}^N[C(\tilde{\mcG}[\gb_i;\theta],\ub_i)].
\end{equation}
Here $C$ denotes a properly defined cost functional which is often taken as the relative mean square error. 

\subsection{Gradient-Based Meta-Learning Methods}

One of highly successful meta-learning algorithms is Model Agnostic Meta-Learning (MAML) \citep{finn2017model}, which led to the development of a series of related gradient-based meta-learning (GBML) methods \citep{raghu2019rapid,nichol2018reptile,antoniou2018train,hospedales2020meta}. 
%
Almost-No-Inner-Loop algorithm (ANIL) \citep{raghu2019rapid} modifies MAML  by freezing the final layer representation during local adaptation. 
Recently, theoretical analysis~\citep{collins2022maml} 
found that the driving force causing MAML and ANIL to recover the general representation is the adaptation of the final layer of their models, which harnesses the underlying task diversity to improve the representation in all directions of interest.

Beyond applications such as image classification and reinforcement learning, a few meta-learning approaches  have studied  hidden-physics learning under meta~\citep{mai2021use,zhang2022metanor,yin2021leads,wang2021meta} or even transfer setting~\citep{kailkhura2019reliable,goswami2022deep}.  Among these meta-learning works, \citep{mai2021use,zhang2022metanor} are designed for specific physical applications, while \citep{yin2021leads,wang2021meta} focus on on dynamics forecasting by learning the temporal evolution information directly~\citep{yin2021leads} or learning time-invariant features~\citep{wang2021meta}. Hence, none of these works have provided a generic approach nor theoretical understanding on how to transfer the multi-task knowledge between a series of complex physical systems, such that all these tasks are governed by a common parametric PDE with different physical parameters.


\section{Meta-Learnt Neural Operator}

To transfer the multi-task knowledge between a series of complex systems governed by different hidden physical parameters, we proposed to leverage the integral neural operator with a meta-learning setting. Before elaborating 
our novel meta-learnt neural operator architecture, MetaNO, we formally state the transfer-learning problem setting for PDE with different parameters.

Assume that we have a set of training tasks $\{\mcT^\eta\}$ such that $\mcT^\eta\sim p(\mcT)$, and for each training task we have a set of observations of loading field/respond field data pairs $\mcD^\eta:=\{(\gb_i^\eta(\xb),\ub^\eta_{i}(\xb))\}_{i=1}^{N^{\eta}}$. Each task can be modeled with a parametric PDE form
\begin{equation}\label{eqn:pde}
\begin{aligned} 
\mcK_{\bb^\eta(\xb)} [\ub_i^\eta] (\xb)=\gb_i^\eta (\xb),\quad&\xb\in \omg,
\end{aligned}
\end{equation}
where $\bb^\eta(\xb)$ is the hidden task-specific physical parameter field for the common governing law. Given a new and unseen test task, $\mcT^{test}$, and a (usually small) context set of labelled samples $\mcD^{\text{test}}:=\{(\gb_i^{\text{test}}(\xb),\ub^{\text{test}}_{i}(\xb))\}_{i=1}^{N^{\text{test}}}$ on it, our goal is to obtain the approximated solution operator model on the test task as $\tilde{\mcG}[\gb;\theta^{\text{test}}]$. To provide a quantitative metric of the performance for each method, we reserve a separate set of labelled samples on the test task as the target set, and measure  averaged relative errors of $\ub$ on this set. In the few-shot learning context, we are particularly interested in the small-sample scenario where $N^{\text{test}}\ll N^{\eta}$. 

\subsection{A Novel Meta-Learnt Neural Operator Architecture}\label{sec:metano}

\begin{algorithm}[ht!]
\caption{MetaNO 
}\label{alg:main}
\begin{algorithmic}
\STATE {\textbf{Meta-Train Phase:}}
\STATE {\textbf{Input:} a batch $\{\mcT^\eta\}_{\eta=1}^H$ of training tasks and labelled data pairs $\mcD^\eta:=\{(\gb_i^\eta(\xb),\ub^\eta_{i}(\xb))\}_{i=1}^{N^{\eta}}$ on each task.}
\STATE {\textbf{Output:} common parameters $\theta^*_I$ and $\theta^*_Q$ across all tasks.}
\STATE {1. Initialize $\theta_I$, $\theta_Q$, and $\{\theta^\eta_P\}_{\eta=1}^H$.}
\STATE {2. Solve for $[\{\theta^{\eta,*}_P\}_{\eta=1}^H,\theta^*_I,\theta^*_Q]$ from the optimization problem in \eqref{eq:train}.}
\\\hrulefill
\STATE {\textbf{Meta-Test Phase:}}
\STATE {\textbf{Input:} a test task $\mcT^{\text{test}}$ and few labelled data pairs $\mcD^{\text{test}}:=\{(\gb_i^{\text{test}}(\xb),\ub^{\text{test}}_{i}(\xb))\}_{i=1}^{N^{\text{test}}}$ on it.}
\STATE {\textbf{Output:} the task-wise parameter $\theta_P^{\text{test},*}$ and the corresponding surrogate PDE solution operator $\tilde{\mcG}[\gb;[\theta_P^{\text{test},*},\theta^*_I,\theta^*_Q]](\xb)$ for the test task.}
\STATE {3. Solve for the lift layer parameter $\theta^{\text{test},*}_P$ from the optimization problem in \eqref{eq:test}.
}
\STATE {4.  \textit{(For cases with large $N^{\text{test}}$ and/or small $N^\eta$)}, fine tune all parameters on the test task.}
\end{algorithmic}
\end{algorithm}
We now propose MetaNO, which applies \textit{task-wise adaptation only to the first layer, i.e., the lifting layer}, with the full algorithm outlined in Algorithm \ref{alg:main}. We point out that MetaNO is substantially different from existed popular meta-learning approaches such as MAML and ANIL, since the later rely on the adaptation of their last layer, as shown in \cite{collins2022maml}. This property makes MetaNO more suitable for PDE solving tasks as will be discussed in theoretical analysis below and confirmed in empirical evaluations of Section \ref{sec:test}.

Similar as in other meta-learning approaches~\citep{yoon2018bayesian,vanschoren2018meta,yang2022efficient,kalais2022stochastic}, the MetaNO algorithm consists of two phases: $1)$  a meta-train phase which learns shared iterative layers parameters $\theta_I$ and projection layer parameters $\theta_P$ from training tasks; $2)$  a meta-test phase which transfers the learned knowledge and rapidly learning surrogate solution operators for unseen test tasks with unknown physical parameter field, where only a few labelled samples are provided. In the meta-train phase, a batch $\{\mcT^\eta\}_{\eta=1}^H$ of $H$ tasks is drawn from the training tasks set, with a context set of $N^\eta$ numbers of labelled loading field/response field data pairs, $\mcD^\eta:=\{(\gb_i^\eta(\xb),\ub^\eta_{i}(\xb))\}_{i=1}^{N^{\eta}}$, provided on each task. Then, we seek the common iterative ($\theta_I$) and projection ($\theta_Q$) parameters, and the task-wise lifting parameters $\theta^\eta_P$ by solving the optimization problem:
\begin{equation}\label{eq:train}
[\{\theta^{\eta,*}_P\}_{\eta=1}^H,\theta^*_I,\theta^*_Q]=\underset{\{\{\theta^\eta_P\}_{\eta=1}^H,\theta_I,\theta_Q\}}{\text{argmin}}\sum_{\eta=1}^H\mcL_{\mcD^\eta}([\theta_P^{\eta},\theta_I,\theta_Q]).
\end{equation}
Then, in the meta-test phase, we adapt the knowledge to a new and unseen test task $\mcT^{\text{test}}$, with limited data on the context set $\mcD^{\text{test}}:=\{(\gb_i^{\text{test}}(\xb),\ub^{\text{test}}_{i}(\xb))\}_{i=1}^{N^{\text{test}}}$ on this task. In particular, we fix the common parameters $\theta^*_I$ and $\theta^*_Q$, then solve for the task-wise parameter $\theta^{\text{test}}_P$ via:

\begin{equation}\label{eq:test}
\theta^{\text{test},*}_P=\underset{\theta^{\text{test}}_P}{\text{argmin}}\;\mcL_{\mcD^{\text{test}}}([\theta_P^{\text{test}},\theta^*_I,\theta^*_Q]).
\end{equation}
One can then fine tune all test task parameters $[\theta_P^{\text{test}},\theta_I,\theta_Q]$ for  further improvements. Finally, the surrogate PDE solution operator on the test task is obtained as:

$$\tilde{\mcG}[\gb;[\theta^{\text{test},*}_P,\theta^*_I,\theta^*_Q]](\xb):=\mathcal{Q}_{\theta^*_Q}\circ(\mathcal{J}_{\theta^*_I})^L\circ\mathcal{P}_{\theta^{\text{test},*}_P}[\gb](\xb).$$

and will be evaluated on a reserved target data set on the test task.

\subsection{Universal Solution Operator}

To see the inspiration of the proposed architecture, without loss of generality, we assume that the underlying task parameter field $\bb^\eta(\xb)$, modeling the physical property field, is normalized and satisfying $\vertii{\bb^\eta(\xb)-\overline{\bb}(\xb)}_{L^2(\Omega)}\leq 1$ for all $\eta\in\{1,\cdots,H\}$, where $\overline{\bb}:=\mathbb{E}_{\mcT^\eta\sim p(\mcT)} [\bb^\eta]$. Denoting $\mcF_{\ub}[\bb]:=\mcK_{\bb}[\ub]$ as a function from physical parameter fields $\mcB$ to loading fields $\mcA$, we take the Fr\'echet derivative of $\mcF$ with respect to $\bb-\overline{\bb}$ and obtain:
$$\mcK_{\bb^\eta}[\ub]=\mcF_{\ub}[{\overline{\bb}}]+D\mcF_{\ub}[\overline{\bb}] (\bb^\eta-\overline{\bb})+o(\vertii{\bb^\eta-\overline{\bb}}_{L^2(\Omega)}).$$
Substituting the above formulation into \eqref{eqn:pde} yields:
$$\mcF_{\ub_i^\eta}[{\overline{\bb}}]+D\mcF_{\ub_i^\eta}[\overline{\bb}] (\bb^\eta-\overline{\bb})\approx \gb_i^\eta.$$
Denoting $\Fb_1[\bb^\eta]:=[\mathbf{1},\bb^\eta-\overline{\bb}]$ and $\Fb_2[\ub_i^\eta]:=[\mcF_{\ub_i^\eta}[{\overline{\bb}}],D\mcF_{\ub_i^\eta}[\overline{\bb}]]$, we can reformulate \eqref{eqn:pde} into a more generic form:
\begin{equation}\label{eqn:pde_1}
\begin{aligned} 
\Fb_1[\bb^\eta](\xb) \cdot \Fb_2[\ub_i^\eta] (\xb)=\gb_i^\eta (\xb),\quad&\xb\in \omg.
\end{aligned}
\end{equation}
Note that this parametric PDE form is very general and applicable to many science and engineering applications --  besides our motivating example on material modeling, other examples  include the monitoring of tissue degeneration problems \citep{zhang2017modeling}, the detection of subsurface flows \citep{dejam2017pre}, the nondestructive inspection in aviation \citep{fallah2019computational}, and the prediction of concrete structures deterioration \citep{wei2021hydro}, etc.

In the following, we show that MetaNOs are universal solution operators for the multi-task PDE solving problem in  \eqref{eqn:pde_1}, in the sense that they can approximate a fixed point method to a desired accuracy. For simplicity, we consider a $1D$ domain $\Omega\subset\real$, and scalar-valued functions $\Fb_1[\bb^\eta]$, $\Fb_2[\ub_i^\eta]$. These functions are assumed to be sufficiently smooth and measured at uniformly distributed nodes $\chi :=\{\xb_1,\xb_2,\dots,\xb_M\}$, with $\Fb_1[\bb^\eta](\xb_j)\neq 0$ for all $\eta$ and $j$. Then, \eqref{eqn:pde_1} can be formulated as an implicit system of equations: 
\begin{equation}\label{eqn:R}
\mcH(\Ub_i^{\eta,*};\tilde{\Gb}_i^{\eta}):=\left[\begin{array}{c}
\Fb_2[\ub_i^\eta] (\xb_1)-\gb_i^\eta (\xb_1)/\Fb_1[\bb^\eta](\xb_1)\\
\vdots\\
\Fb_2[\ub_i^\eta] (\xb_M)-\gb_i^\eta (\xb_M)/\Fb_1[\bb^\eta](\xb_M)\\
\end{array}\right]=\mathbf{0},
\end{equation}
where $\Ub_i^{\eta,*} := 
[\ub_i^\eta(\xb_1),\dots,\ub_i^\eta(\xb_M)]$ is the solution we seek, 
$\tilde{\Gb}_i^{\eta} := [\gb_i^\eta(\xb_1)/\Fb_1[\bb^\eta](\xb_1),\dots,\gb_i^\eta(\xb_M)/\Fb_1[\bb^\eta](\xb_M)]$
is the reparameterized loading vector, 
and $\Gb_i^{\eta} := [\gb_i^\eta(\xb_1),\gb_i^\eta(\xb_2),\dots,\gb_i^\eta(\xb_M)]$ is the original loading vector. Here, we notice that all task-specific information is encoded in $\tilde{\Gb}_i^{\eta}$ and can be captured in the lifting layer parameter. Therefore, when seeing \eqref{eqn:R} as an implicit problem of $\Ub_i^{\eta,*}$ and $\tilde{\Gb}_i^{\eta}$, it is actually independent of the task parameter field $\bb^\eta$, i.e., this problem is task-independent. In the following, we refer to \eqref{eqn:R} without the task index, as $\mcH(\Ub^*;\tilde{\Gb})$, for notation simplicity. 

To solve for $\Ub^*$ from the nonlinear system $\mcH(\Ub^*;{\tilde{\Gb}})=\mathbf{0}$, a popular approach would be to use fixed-point iteration methods such as the Newton-Raphson method. With an initial guess of the solution (denoted as $\Ub^0$), the process is repeated to produce successively better approximations to the roots of \eqref{eqn:R}, from the solution of iteration $l$ (denoted as $\Ub^l$) to that of $l+1$ (denoted as $\Ub^{l+1}$) as:
\begin{equation}\label{eqn:newton}
 \Ub^{l+1}=\Ub^l-(\nabla \mcH(\Ub^l;\tilde{\Gb}))^{-1}\mcH(\Ub^l;\tilde{\Gb}):=\Ub^l+\mcR(\Ub^l,\tilde{\Gb}),
\end{equation}
until a sufficiently precise value is reached. In the following, we show that as long as Assumptions~\ref{asp:1} and ~\ref{asp:2}  hold, i.e., there exists a converging fixed point method
, then MetaNO can be seen as an resemblance of the fixed point method in \eqref{eqn:newton} and hence acts as an universal approximator of the solution operator for \eqref{eqn:pde_1}.
\begin{assumption}\label{asp:1}
There exists a fixed point equation, $\Ub=\Ub+\mcR(\Ub,\tilde{\Gb})$ for the implicit of problem \eqref{eqn:R}, such that $\mathcal{R}:\mathbb{R}^{2M} \mapsto \mathbb{R}^M$ is a continuous function satisfying $\mathcal{R}(\Ub,\tilde{\Gb}) = \mathbf{0}$ and $||\mathcal{R}(\hat{\Ub},\tilde{\Gb})-\mathcal{R}(\tilde{\Ub},\tilde{\Gb})||_{l^2(\mathbb{R}^M)} \leq m ||\hat{\Ub}-\tilde{\Ub}||_{l^2(\mathbb{R}^M)}$ for any two vectors $\hat{\Ub},\tilde{\Ub}\in\real^{M}$. Here, $m\geq 0$ is a constant independent of $\tilde{\Gb}$.
   \end{assumption}
  \begin{assumption}\label{asp:2}
   With the initial guess $\Ub^0:=[\xb_1,\cdots,\xb_M]$, the fixed-point iteration $\Ub^{l+1} = \Ub^{l}  + \mathcal{R}(\Ub^l,\tilde{\Gb})$ ($l=0,1,\dots$) converges, i.e., for any given $\varepsilon > 0$, there exists an integer $L$ such that 
    \begin{equation*}
        ||\Ub^l - \Ub^*||_{l^2(\mathbb{R}^M)} \leq \varepsilon, \quad \forall l > L,
    \end{equation*}
for all possible input instances $\tilde{\Gb}\in\real^M$ and their corresponding solutions $\Ub^*$.
\end{assumption}
Intuitively, Assumptions~\ref{asp:1} and ~\ref{asp:2} ensure the hidden PDEs to be numerically solvable with a converging iterative solver, which is a typical required condition of numerical PDE solving problems. 
Then, we have our universal approximation theorem as below, with proof provided in \ref{sec:theorem1}. The main result of this theorem is to show that for any desired accuracy $\varepsilon>0$, one can find a sufficiently large $L>0$ and sets of parameters $\theta^\eta=\{\theta^\eta_P,\theta_I,\theta_Q\}$, such that the resultant MetaNO model acts as a fixed point method with the desired prediction 
for all tasks and samples.
\begin{theorem}[Universal approximation] \label{thm:main}
Given Assumptions \ref{asp:1}-\ref{asp:2}, let the activation function $\sigma$ for all iterative kernel integration layers be the ReLU function, and the activation function in the projection layer be the identity function. Then for any $\varepsilon > 0$, there exist sufficiently large layer number $L>0$ and feature dimension number $d_h>0$, such that one can find a parameter set for the multi-task problem, $\theta^\eta=[\theta^\eta_P,\theta_I,\theta_Q]$, such that the corresponding MetaNO model satisfies 
\begin{equation*}
    \vertii{\mathcal{Q}_{\theta_Q}\circ(\mathcal{J}_{\theta_I})^L\circ \mathcal{P}_{\theta^\eta_P}([\Ub^0, \Gb^\eta ]^{\mathrm{T}})-\Ub^{\eta,*}}\leq\varepsilon,
\end{equation*}
for all loading instance $\Gb^\eta\in\real^M$ and tasks.
\end{theorem}

\begin{figure*}
    \centering
    \includegraphics[width=0.99\textwidth]{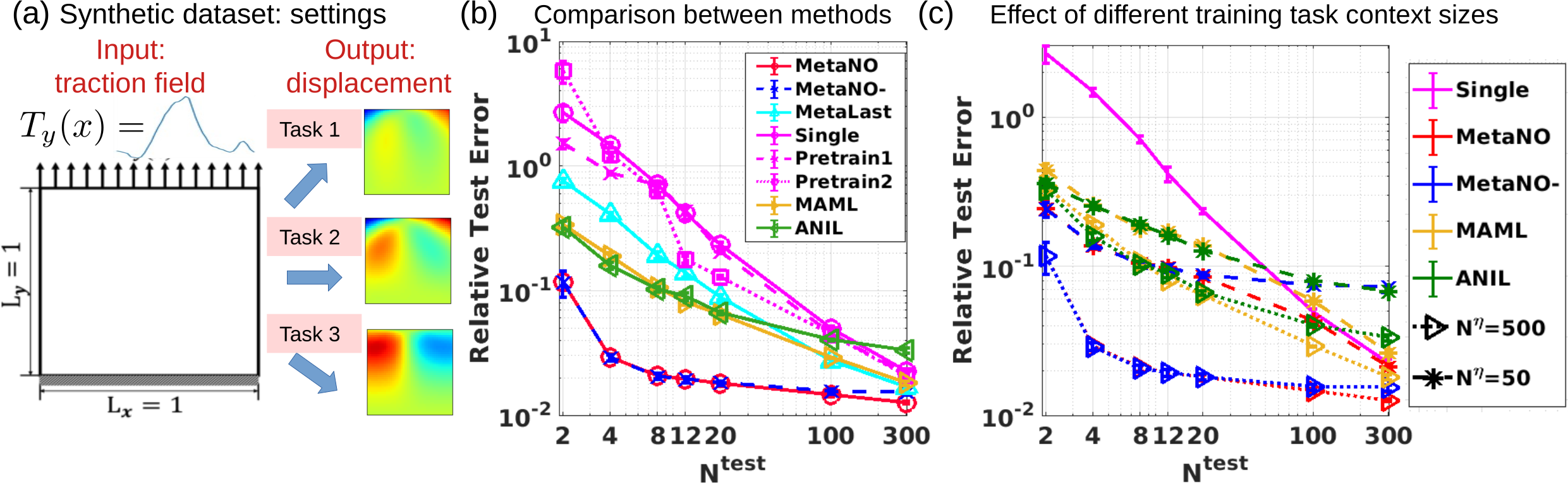}
    \caption{Results on the synthetic data set. (a) The problem setting and visualization of the ground-truth solution $u_x^\eta(\xb)$ from different tasks, showing the solution diversity across tasks due to the change of underlying parameter set $\bb^\eta$. (b) The ablation study comparison on test errors in the in-distribution test, when using the full context set ($N^\eta=500$) on training tasks and different sizes of context set ($N^{\text{test}}$) on test tasks. (c) The ablation study showing the effect of varying training task context set sizes. More results can be found in \ref{sec:ablation}.}
    \label{fig:metap_hgo}
\end{figure*}

\section{Empirical Evaluation}\label{sec:test}

In this section, we demonstrate the empirical effectiveness of the proposed MetaNO approach. Specifically, we conduct experiments on a synthetic dataset from a nonlinear PDE solving problem, a benchmark dataset of heterogeneous materials subject to large deformation, and a real-world dataset from biological tissue mechanical testing. We compare the proposed method against competitive GBML methods {as well as two non-meta transfer-learning baselines}. All of the experiments are implemented using PyTorch with Adam optimizer, with a brief description of each method provided in the \ref{sec:expdetails}. In all experiments, we considered the averaged relative error, $||\ub_{i,pred}-\ub_{i}||_{L^2(\omg)}/||\ub_{i}||_{L^2(\omg)}$, as the error metric. We repeat each experiment for 5 times, and report the averaged relative errors and their standard errors.

\subsection{Synthetic Data Sets and Ablation Study}

We first consider the PDE-solution-finding problem of the Holzapfel-Gasser-Odgen (HGO) model~\citep{holzapfel2000new}, which describes the deformation of hyperelastic, anisotropic, and fiber-reinforced materials. Different tasks correspond to different material parameter sets $\{k_1,k_2,E,\nu,\alpha \}$, where $k_{1}$ and $k_{2}$ are fiber modulus and the exponential coefficients, respectively, $E$ is the Young's modulus, $\nu$ is the Poisson ratio, and $\alpha$ is the fiber angle direction from the reference direction. 
The physical response of interest is the displacement field $\ub:[0,1]^2\rightarrow\real^2$ , subject to different traction loadings applied on the top edge of this material. Therefore, we take the input function $\gb(\xb)$ as the padded traction loading field, and the output function as the corresponding displacement field. We provide more detailed discussions on data generation process and hyperparameters used by each method in \ref{sec:expdetails}.

To investigate the performance of MetaNO in few-shot learning, we generate 59 training, 1 validation tasks, and 5 in-distribution (ID) test tasks by sampling different physical parameters $k_1, k_2, E, \nu, \alpha$ from the same uniform distribution. 
To further evaluate the generalizability when the physical parameters of test tasks are outside the training regime, we also generate 2 out-of-distribution (OOD) test tasks with physical parameters from different distributions. 
The distribution of training and ID/OOD tasks are demonstrated in Figure \ref{fig:distri} of \ref{sec:expdetails}, where one can see that the first OOD task (denoted as ``OOD Task1'') corresponds to a stiffer material sample and smaller deformation for each given loading, while the second OOD task (denoted as ``OOD Task2'') generates a softer material sample and larger deformation. For each training task, we generate 500 data pairs $\mcD^\eta:=\{(\gb_i^\eta,\ub_i^\eta)\}_{i=1}^{500}$, by sampling the vertical traction loading from a Gaussian random field. 
Then, the corresponding ground-truth displacement field is obtained using the finite element method implemented in FEniCS \citep{alnaes2015fenics}. For test tasks, we train with $N^{\text{test}}=\{2,4,8,12,20,100,300\}$ numbers of labelled data pairs (the context set), and evaluate the model on a reserved dataset with $200$ data pairs (the target set) on each test task. An $8$-layer IFNO is employed as the base model. 

\textbf{Ablation Study.} We first conduct an ablation study on 3 variants of the proposed algorithm: 
 1) to use the full meta-train and meta-test phases as in Algorithm \ref{alg:main} (denotes as ``MetaNO''); 2) to perform steps 1-3 of Algorithm \ref{alg:main}, such that only the lifting layer is adapted in the meta-test phase (denotes as ``MetaNO-''); 
3) to apply task-wise adaptation only to the projection layer instead of the lift layer in both meta-train and meta-test phases (denoted as ``MetaLast'').
We study if the successful ``adapting last layers'' strategy of MAML and ANIL in image classification problems would apply for our PDE solving problem. Besides these three settings, we also report the few-shot learning results with five baseline methods: 1) Learn a neural operator model only based on the context data set of the test task (denoted as ``Single''); 2) Pretrain a  neural operator model based on all training task data sets, then fine-tune it based on the context test task data set (denoted as ``Pretrain1''); 3) Pretrain a single neural operator model based on the context data set of one training task, then fine-tune it based on the context test task data set (denoted as ``Pretrain2''); To remove the possible dependency on the pre-training task, in this baseline we randomly select five training tasks for the purpose of pretraining and report the averaged results. 4) MAML, and 5) ANIL. For all experiments we use the full context data set on each training task ($N^\eta=500$). As shown in Figure \ref{fig:metap_hgo}(b), MetaNO- and MetaNO are both able to quickly adapt with few data pairs -- to achieve a test error below $5\%$, ``Single'' and the {two transfer-learning baselines (``Pretrain1'', ``Pretrain2'')} require 100+ data pairs, while MetaNO- and MetaNO requires only 4 data pairs. On the other hand, MetaLast, MAML and ANIL have similar performance. They 
all require 100 data pairs to achieve a $<5\%$ test error. This observation verifies our finding on the multi-task parametric PDE solution operator learning problem, where one should adapt the first layer, not the last ones. Moreover, when comparing MetaNO- and MetaNO, we can see that the additional fine-tune step  improves the performance in the larger-sample regime (when $N^{\text{test}}\geq 100$). This fact shows that when given sufficient training context sets,  adapting the first layer can capture the underlying task diversity  so  further fine-tuning may not be  needed.

\textbf{Effect of Varying Training Context Set Sizes. }In this study, we investigate the effect of different training task context sizes $N^\eta=\{50,100,200,500\}$ on four meta-learnt models: MetaNO, MetaNO-, MAML, and ANIL. Due to the limit of space, in Figure \ref{fig:metap_hgo}(c) we demonstrate the efficacy of each method when using the largest training context set ($N^\eta=500$) and the smallest training context set ($N^\eta=50$), and leave further results (see top Figure \ref{fig:metap_in_out}) and discussions in \ref{sec:ablation}. One can see that when $N^{\text{test}}\leq 20$, MetaNO- and MetaNO have similar performance and consistently beat MAML and ANIL for both context set sizes. With the increase of $N^{\text{test}}$, the fine-tuning strategy on the test context set becomes more helpful where we see MetaNO becomes more accurate than MetaNO- and MAML beats ANIL. Such effect is more evident on small training context set cases. In all combinations of $N^{\eta}$ and $N^{\text{test}}$, MetaNO achieves the best performance among all models.

\textbf{In-Distribution and Out-Of-Distribution Tests.} On bottom Figure \ref{fig:metap_in_out} in \ref{sec:ablation}, we demonstrate the relative test error of MetaNO against MAML in both ID and OOD tasks. 
We can see that test errors of these 3 tasks are  in a similar scale as the error on training tasks. 
In all three cases, MetaNO outperforms MAML, 
hence validating the good generalization performance of MetaNO. For more discussion, please refer to \ref{sec:ablation}.

\begin{figure*}
\centering%
\includegraphics[width=1\textwidth]{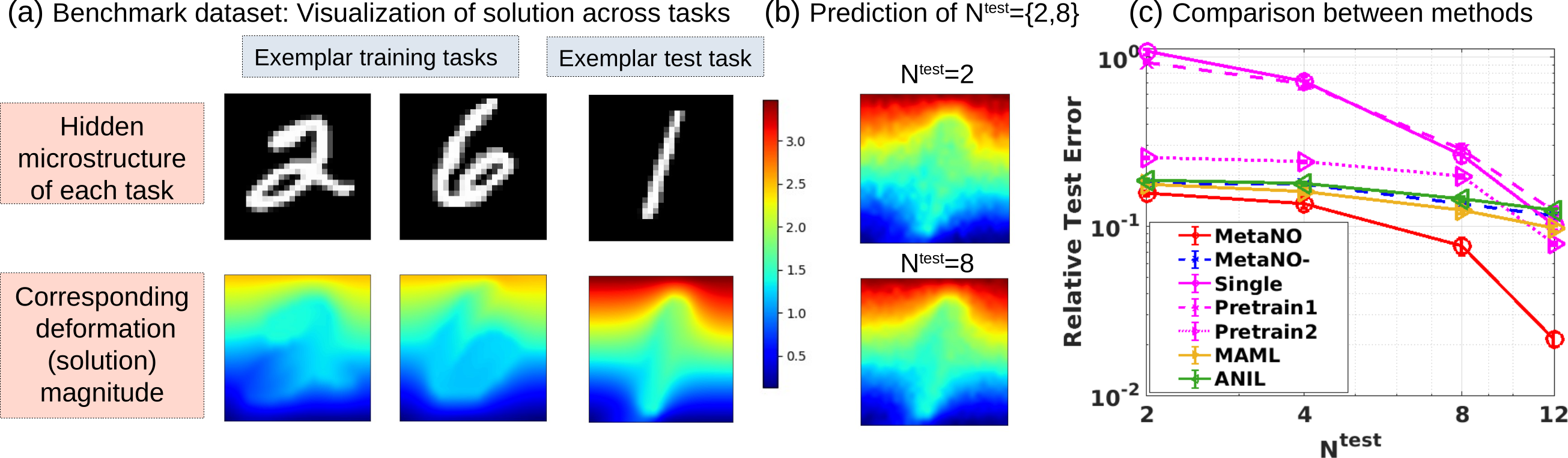}

\caption{Results on the benchmark (Mechanical MNIST \cite{lejeune2020mechanical}) dataset. (a) The visualization of different tasks, their underlying microstructure field $\bb^\eta$, and the corresponding ground-truth solution. (b) Prediction results based on few samples ($N^{\text{test}}=2$ and $N^{\text{test}}=8$) on a test task. (c) Comparison of MetaNO and five baseline methods. 
}%
\label{fig:MMNIST}
\end{figure*}

\subsection{Benchmark Mechanical MNIST Datasets}

We further test MetaNO and five baseline methods on  benchmark Mechanical MNIST \citep{lejeune2020mechanical}. Mechanical MNIST is a dataset of heterogeneous material undergoing large deformation. It contains 70,000 heterogeneous material specimens, and each specimen is governed by the Neo-Hookean material with a varying modulus converted from the MNIST bitmap images. On each specimen, 32 loading/response data pairs are provided\footnote{We have excluded small deformation samples with the maximum displacement magnitude $\leq 0.1$.}. 
Here in, we randomly select one specimen corresponding to hand-written number $0$ and $2-9$ respectively as training tasks. Then, among the specimens corresponding to $1$, we randomly select six specimens: one for validation and the rest five as the test tasks. Visualization of the ground-truth solutions corresponding to one common loading from different tasks is provided in Figure \ref{fig:MMNIST}(a), together with the underlying (hidden) microstructure pattern which determines the parameter set $\bb^\eta$. On the meta-train phase, we use the full context data set of all $32$ samples for each training task. On the meta-test phase, we reserve $20$ data pairs on the test task as the target set for evaluation, then train each model under the few-shot learning setting with $N^{\text{test}}=\{2,4,8,12\}$ labelled data pairs as the context set. All approaches are developed based on an 32-layer IFNO model.

Besides the diversity of tasks as seen in Figure \ref{fig:MMNIST}(a), notice that we also have a small number of training tasks ($H=9$), and a relatively small training context set size ($N^\eta=32$). All these facts make the transfer learning on this benchmark dataset challenging. 
We present the results in Figure \ref{fig:MMNIST}(b) and (c). The neural operator model learned by MetaNO again outperforms the baseline {single/transfer learning} models and the state-of-the-art GBML models. Our MetaNO model achieves $15\%$ error when using only 2 labelled data pair on the test task, while the Single model has high errors due to overfitting. This fact highlights the importance of learning across multi-tasks:  when the total number of measurements on each specimen is limited, it is necessary to transfer the knowledge across specimens. Moreover, while MetaNO-, MAML, and ANIL all have a similar performance in this example , the fine-tuning step in MetaNO seems to  substantially improve the accuracy, especially when $N^{\text{test}}$ gets larger. This observation is consistent with previous finding on varying training task context sizes. 


\begin{figure}
\centering%
\includegraphics[width=0.4\columnwidth]{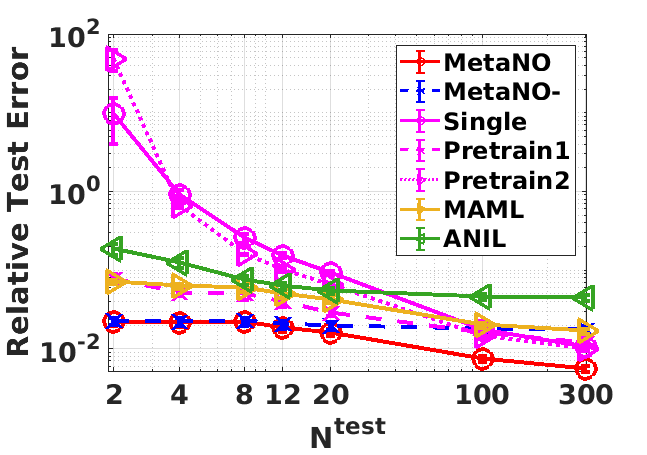}
\caption{Results on the real-world dataset (heart valve tissue), which features measurement noise and a small number of available tasks. Comparison of MetaNO and five baseline methods.}%
\label{fig:tissue}
\end{figure}

\subsection{Application on Real-World Data Sets}

We now take a step further to demonstrate the performance of our method on a real-world physical response dataset, which is not generated by solving PDEs. We consider the problem of learning the mechanical response of multiple biological tissue specimens from DIC displacement tracking measurements. As demonstrated in Figure \ref{fig:domain}, we measure the biaxial loading of tricuspid valve anterior leaflet (TVAL) specimens from a porcine heart, such that each specimen (as a task) corresponds to a different region of the leaflet. Due to  material heterogeneity of biological tissues, these specimens contain  different mechanical and structural properties. 

In this experiment, we aim to model the tissue response by learning a neural operator mapping the boundary displacement loading to the interior displacement field on each tissue specimen. On each specimen, we have 500 available data pairs. Due to expenses of obtaining the experimental tissue, only 16 specimens are available in total.  
This reflects a common challenge in scientific applications, we not only have limited samples per task, the number of available training tasks is also limited. In the experiment, we use 13 specimens for training and validation with context size $N^\eta=500$, and provide the test results as the average on the rest 3 specimens. With a 4-layer IFNO as the base model, we train each model based on $N^{\text{test}}\in[2,300]$ 
samples, and then evaluate the performance on another $200$ samples. The results are provided in Figure \ref{fig:tissue}. MetaNO performs the best among all the methods across all $N^{\text{test}}$,  beating  MAML and ANIL by a significant margin. 
Interestingly, MAML and ANIL did not even beat the ``Pretrain1'' method, possibly due to the low efficacy of the adapting last layers strategy and the small number of training tasks. 

\section{Conclusion}
In this paper we propose MetaNO, the first neural-operator-based meta-learning approach that are designed to achieve good transferability in learning complex physical system responses. Our MetaNO features a novel \textit{first layer adaption architecture}, which is theoretically motivated and shown to be the universal solution operator for multiple parametric PDE solving tasks. We demonstrate the effectiveness of our proposed MetaNO algorithm on various synthetic, benchmark, and real-world datasets, showing promises with significant improvement in sample efficiency over baseline methods. For future work, we will investigate the applicability of the proposed approach to other scientific domains.




\bibliographystyle{elsarticle-num}

\bibliography{yyu}


\appendix


\section{Proof of Theorem 1}
\label{sec:theorem1}

In this section we provide the detailed proof for Theorem 1, based on Assumptions~\ref{asp:1} and \ref{asp:2}. Intuitively, these assumptions mean the underlying implicit problem is solvable with a converging fixed point method. This condition is a basic requirement by numerical PDEs, and it generally holds true in many applications governed by nonlinear and complex PDEs, such as in our three experiments.

Here, we prove that the MetaNO is universal, i.e., given a fixed point method satisfying  Assumptions \ref{asp:1} and \ref{asp:2}, one can find parameter sets $\theta^\eta$ whose output approximates $\Ub^{\eta,*}$ to a desired accuracy, $\varepsilon>0$, for all $\eta=1,\cdots,H$ tasks. For the task-wise parameters, with a slight abuse of notation, we denote $P^\eta\in\real^{d_hM\times (d_g+s)M}$ as the collection of the pointwise weight matrices at each discretization point in $\chi$ for the $\eta$-th task, and $\pb^\eta\in\real^{d_hM}$ for the bias in the lifting layer. Then, for the parameters shared among all tasks, in the iterative layer we denote $\Cb=[\cb(\xb_1),\cdots,\cb(\xb_M)]\in\real^{d_hM}$ as the collection of pointwise bias vectors $\cb(\xb_i)$, $W\in\real^{d_h\times d_h}$ for the local linear transformation, and $R=\mathcal{F}[\kappa(\cdot;\vb)]\in\complex^{d_h\times d_h\times M}\in\complex^{d_h\times d_h\times M}$ for the Fourier coefficients of the kernel $\kappa$. For simplicity, here we have assumed that the Fourier coefficient is not truncated, and all available frequencies are used. Then, for the projection layer we seek  $Q_1\in\real^{d_QM\times d_hM}$, $Q_2\in\real^{d_u M\times d_QM}$, $\qb_1\in\real^{d_QM}$ and $\qb_2\in\real^{d_uM}$. For the simplicity of notation, in this section we organize the feature vector $\Hb\in\real^{d_hM}$ in a way such that the components corresponding to each discretization point are adjacent, i.e., $\Hb=[\Hb(\xb_1),\cdots,\Hb(\xb_M)]$ and $\Hb(\xb_i)\in\real^{d_h}$. 

We point out that under this circumstance, the (discretized) iterative layer can be written as
\begin{align*}
\mcJ[\Hb(l)]=&\Hb(l)+ \dfrac{1}{L}\sigma\left(\tilde{W}\Hb(l)+\text{Re}(\mathcal{F}_{\Delta x}^{-1}(R\cdot \mathcal{F}_{\Delta x}(\Hb(l))))+\Cb\right)\\
=&\Hb(l)+ \dfrac{1}{L}\sigma\left(V\Hb(l)+\Cb\right),
\end{align*}
with 
{$$ V := \text{Re}\begin{bmatrix}
    \sum\limits_{n=0}^{M-1} R_{n+1}+W & \sum\limits_{n=0}^{M-1} R_{n+1} \exp(\frac{2i\pi\Delta x n}{M}) &
    \dots & \sum\limits_{n=0}^{M-1} R_{n+1} \exp(\frac{2i\pi(M-1)\Delta x n}{M}) \\
     \sum\limits_{n=0}^{M-1} R_{n+1} \exp(\frac{2i\pi\Delta x n}{M}) 
     & \sum\limits_{n=0}^{M-1} R_{n+1}+W & \dots  & \sum\limits_{n=0}^{M-1} R_{n+1} \exp(\frac{2i\pi(M-2)\Delta x n}{M}) \\
     \vdots & \vdots & \ddots & \vdots \\
     \sum\limits_{n=0}^{M-1} R_{n+1} \exp(\frac{2i\pi(M-1)\Delta x n}{M})
      & \sum\limits_{n=0}^{M-1} R_{n+1} \exp(\frac{2i\pi(M-2)\Delta x n}{M}) &\dots&  \sum\limits_{n=0}^{M-1} R_{n+1}+W \\
    \end{bmatrix}.$$}
Here, $R\in\complex^{M\times d_h\times d_h}$ with $R_i\in\complex^{d_h\times d_h}$ being the component associated with each discretization point $\xb_i\in\chi$, ${V}\in\real^{d_hM\times d_hM}$, $\Cb\in\real^{d_hM}$, $\tilde{W}:=W\oplus W\oplus \cdots \oplus W$ is a $d_hM\times d_hM$ block diagonal matrix formed by $W\in\real^{d_h\times d_h}$,  $\mathcal{F}_{\Delta x}$ and $\mathcal{F}_{\Delta x}^{-1}$ denote the discrete Fourier transform and its inverse, respectively. By further taking $R_2=\cdots=R_M=W=0$, a $d_h\times d_h$ matrix with all its elements being zero, it suffices to show the universal approximation property for an iterative layer as follows:
$$\mcJ(\Hb(l)):=\Hb(l)+ \dfrac{1}{L}\sigma\left(\tilde{V}\Hb(l)+\Cb\right)$$
where $\tilde{V}:=\mathbf{1}_{[M,M]}\otimes V$ with $V\in\real^{d_h\times d_h}$ and $\mathbf{1}_{[m,n]}$ being an $m$ by $n$ all-ones matrix.

To be more precise, we will prove the following theorem:
\begin{thmn}[\textbf{\ref{thm:main}} \textnormal{(Universal approximation)}]
Let $\Ub^{\eta,*} = [\ub^\eta(\xb_1),\ub^\eta(\xb_2),\dots,\ub^\eta(\xb_M)]$ 
be the ground-truth solution of $\eta$-th task that satisfies Assumptions \ref{asp:1}-\ref{asp:2}, the activation function $\sigma$ for all iterative kernel integration layers be the ReLU function, and the activation function in the projection layer be the identity function. Then for any $\varepsilon > 0$, there exist a sufficiently large layer number $L>0$ and feature dimension number $d_h>0$, such that one can find a parameter set for the multi-task problem, $\theta^\eta = [\theta_P^{\eta},\theta_I, \theta_Q]$ with the corresponding MetaNO model satisfies 
\begin{equation*}
    \vertii{\mathcal{Q}_{\theta_Q}\circ(\mathcal{J}_{\theta_I})^L\circ \mathcal{P}_{\theta_P^{\eta}}([\Ub^0, \Gb^{\eta} ]^{\mathrm{T}})-\Ub^{\eta,*}}\leq\varepsilon,\quad \forall \Gb^{\eta}\in\real^M. 
\end{equation*}
\end{thmn}

For the proof of this main theorem, we need the following approximation property of a shallow neural network, with its detailed proof provided in \cite{you2022learning}:
\begin{lemma}\label{lemma:1}
Given a continuous function ${\mathcal{T}}: 
\mathbb{R}^{2M} \mapsto \mathbb{R}^M$, and a non-polynomial and continuous activation function $\sigma$, for any constant ${\hat{\varepsilon}}>0$ there exists a shallow neural network model $\hat{\mathcal{T}}:= S\sigma \left(B\Xb +A \right)$ such that 
\begin{equation*} 
    ||\mathcal{T}(\Xb) - \hat{\mathcal{T}}(\Xb)||_{l^2(\mathbb{R}^{M})} \leq {\hat{\varepsilon}}, \quad \forall \Xb\in\real^{2M},
\end{equation*}
for sufficiently large feature dimension ${\hat{d}}>0$. 
Here, $S\in \mathbb{R}^{M \times {\hat{d}}M}$, $B \in \mathbb{R}^{{\hat{d}}M \times 2M}$, and $A \in \mathbb{R}^{{\hat{d}}M}$ are matrices/vectors which are independent of $\Xb$.
\end{lemma}

We now proceed to the proof of Theorem \ref{thm:main}:
\begin{proof}
Since all $\Ub^{\eta,*}$ satisfies Assumptions \ref{asp:1}-\ref{asp:2}, for any $\varepsilon > 0$, we first pick a sufficiently large integer $L$ such that the $L$-th layer iteration result of this fixed point formulation satisfies $||\Ub^L - \Ub^{\eta,*}||_{l^2(\mathbb{R}^M)} \leq \frac{\varepsilon}{2}$ for all tasks. {By taking $\hat{\varepsilon}:=\frac{m\varepsilon}{2(1+m)^L}$ in Lemma \ref{lemma:1},  
there exists a sufficiently large feature dimension $\hat{d}$ and one can find $S \in \mathbb{R}^{M \times \hat{d}M}$, $B \in \mathbb{R}^{\hat{d}M \times 2M}$, and $A \in \mathbb{R}^{\hat{d}M}$}, such that $\hat{\mathcal{R}}(\Ub^\eta,\tilde{\Gb}^\eta):=S\sigma(B[\Ub^\eta,\tilde{\Gb}^\eta]^{\mathrm{T}} + A)$ satisfies
\begin{equation*}
     ||\mathcal{R}(\Ub^\eta,\tilde{\Gb}^\eta) -\hat{\mathcal{R}}(\Ub^\eta,\tilde{\Gb}^\eta)||_{l^2(\mathbb{R}^{M})} = ||\mathcal{R}(\Ub^\eta,\tilde{\Gb}^\eta) - S\sigma(B[\Ub^\eta,\tilde{\Gb}^\eta]^{\mathrm{T}} + A) ||_{l^2(\mathbb{R}^M)} \leq {\hat{\varepsilon}=}\frac{m\varepsilon}{2(1+m)^L},
\end{equation*}
where $m$ is the contraction parameter of $\mathcal{R}$, as defined in Assumption \ref{asp:1}.
By this construction, we know that $S$ has independent rows. {Denoting $\tilde{d}:=\hat{d}+1>0$,} there exists the right inverse of $S$, which we denote as $S^{+} \in \mathbb{R}^{(\tilde{d}-1)M \times M}$, such that 
\begin{align*}
    SS^+ = I_M, \quad S^+S  := \tilde{I}_{(\tilde{d}-1)M}, 
\end{align*}
where $I_M$ is the $M$ by $M$ identity matrix, $\tilde{I}_{(\tilde{d}-1)M}$ is a $(\tilde{d}-1)M$ by $(\tilde{d}-1)M$ block matrix with each of its element being either $1$ or $0$. Hence, for any vector $Z\in\real{(\tilde{d}-1)M}$, we have $\sigma(\tilde{I}_{(\tilde{d}-1)M}Z)=\tilde{I}_{(\tilde{d}-1)M}\sigma(Z)$. Moreover, we note that $S$ has a very special structure: from the $((i-1)(\tilde{d}-1)+1)$-th to the $(i(\tilde{d}-1))$-th column of $S$, all nonzero elements are on its $i$-th row. Correspondingly, we can also choose $S^+$ to have a special structure: from the $((i-1)(\tilde{d}-1)+1)$-th to the $(i(\tilde{d}-1))$-th row of $S^+$, all nonzero elements are on its $i$-th column. Hence, when multiplying $S^+$ with $\Ub$, there will be no entanglement between different components of $\Ub$. That means, $S^+$ can be seen as a pointwise weight function. 

We now construct the parameters of MetaNO as follows. In this construction, we choose the feature dimension as $d_h := \tilde{d}M$. With the input $[\Ub^0,\Gb^\eta] \in \mathbb{R}^{2M}$, for the lift layer we set 
$$P^\eta := \mathbf{1}_{[M,1]}\otimes\begin{bmatrix}
S^+ & \mathbf{0}\\
\mathbf{0} & D^\eta\\
    \end{bmatrix}=\underbrace{\begin{bmatrix}
S^+ & \mathbf{0}&S^+ & \mathbf{0} & \cdots & S^+ & \mathbf{0}\\
\mathbf{0} & D^\eta &\mathbf{0} & D^\eta& \cdots&\mathbf{0} & D^\eta\\
    \end{bmatrix}^{\mathrm{T}}}_{\text{repeated for }M \text{ times}}\in \real^{d_hM\times 2M},$$
and $\pb^\eta := \mathbf{0}\in \real^{d_hM}$. Here, $D^\eta:=\text{diag}[1/\Fb_1[\bb^\eta](\xb_1),\cdots,1/\Fb_1[\bb^\eta](\xb_M)]$. As such, the initial layer of feature is then given by 
$$\Hb(0) =P^\eta([\Ub^0,\Gb^\eta]^{\mathrm{T}})=\mathbf{1}_{[M,1]}\otimes[S^+\Ub^0, D^\eta\Gb^\eta ]^{\mathrm{T}}=\mathbf{1}_{[M,1]}\otimes[S^+\Ub^0, \tilde{\Gb}^{\eta} ]^{\mathrm{T}} \in \real^{dM}.$$
Here, we point out that $P^{\eta}$ and $\pb^{\eta}$ can be seen as pointwise weight and bias functions, respectively.

Next we construct the shared iterative layer $\mathcal{J}$, by setting
$$V := 
    \begin{bmatrix}
    \tilde{I}_{(
    \tilde{d}-1)M}B/M\\
    0 \\
    \end{bmatrix}
    \begin{bmatrix}
    LS  & \mathbf{0}\\
    \mathbf{0} & LI_M\\
    \end{bmatrix},\;\tilde{V}:=\mathbf{1}_{[M,M]}\otimes V,\;
    \text{ and }\bm{C} := \mathbf{1}_{[M,1]}\otimes\begin{bmatrix}
    L\tilde{I}_{(\tilde{d}-1)M}A \\
    \mathbf{0} \\
    \end{bmatrix}.$$
Note that $\tilde{V}$ is independent of $\eta$, and falls into the formulation of $V$, by letting $R_1 =V$  
 and $R_2=R_2=\cdots=R_{M} = W=0$.      
For the $l+1$-th layer of feature vector, we then arrive at 
{    \begin{align*}
        \Hb&(l+1) = \Hb(l)+ \dfrac{1}{L}\sigma\left(\tilde{V}\Hb(l)+\Cb\right)\\
    = &\Hb(l) + \left(I_M\otimes\begin{bmatrix}
        S^+S  & \mathbf{0} \\
        \mathbf{0} & I_M \\ 
        \end{bmatrix}\right)\sigma \left(\left(\mathbf{1}_{[M,1]}\otimes
        \begin{bmatrix}
    B/M \\
    \mathbf{0} \\
    \end{bmatrix}\right)
    \left(\mathbf{1}_{[1,M]}\otimes\begin{bmatrix}
        S & \mathbf{0}\\
        \mathbf{0} & I_M\\
        \end{bmatrix}\right) \Hb(l) + \mathbf{1}_{[M,1]}\otimes\begin{bmatrix}
        A \\
        \mathbf{0}\\
        \end{bmatrix} \right),        
    \end{align*}}
where $\Hb(l) = [\hat{\hb}_1^{l},  \hat{\hb}_2^{l},\dots, \hat{\hb}_{2M-1}^{l},  \hat{\hb}_{2M}^{l}]^{\mathrm{T}}$ denotes the (spatially discretized) hidden layer feature at the $l-$th iterative layer of the IFNO. Subsequently, we note that the second part of the feature vector, $\hat{\hb}_{2j}^{l}\in\real^{M}$, satisfies
$$\hat{\hb}_{2j}^{l+1}=\hat{\hb}_{2j}^{l}=\cdots=\hat{\hb}_{2j}^{0}=\tilde{\Gb}^{\eta}, \quad \forall l=0,\cdots,L-1, \forall j = 1,\cdots,M$$
Hence, the first part of the feature vector, $\hat{\hb}_{2j-1}^{l}\in\real^{(\tilde{d}-1)M}$, satisfies the following iterative rule:
$$\hat{\hb}_{2j-1}^{l+1}=\hat{\hb}_{2j-1}^{l}+S^+S\sigma(B[S\hat{\hb}_{2j-1}^{l}, \tilde{\Gb}^{\eta} ]^{\mathrm{T}}+A), \quad \forall l=0,\cdots,L-1, \forall j = 1,\cdots,M,$$
and
$$\hat{\hb}_{1}^{l+1}=\hat{\hb}_{3}^{l+1}=\cdots=\hat{\hb}_{2M-1}^{l+1}.$$
Finally, for the projection layer $\mathcal{Q}$, we set the activation function in the projection layer as the identity function, $Q_1 := I_{d_hM}$ (the identity matrix of size $d_hM$), $Q_2 := [S, \mathbf{0}] \in \real^{M \times d_hM}$, $\qb_1 := \mathbf{0}\in\real^{d_hM}$, and $\qb_2 := \mathbf{0}\in\real^{M}$. Denoting the output $\Ub^\eta:=\mathcal{Q}_{\theta_Q}\circ(\mathcal{J}_{\theta_I})^L\circ \mathcal{P}_{\theta_{P}^{\eta}}([\Ub^0, \Gb^{\eta}]^{\mathrm{T}})$, we now show that $\Ub^\eta$ can approximate $\Ub^{\eta,*}$ with a desired accuracy $\varepsilon$:
{\small\begin{align*}
    ||\Ub^\eta - \Ub^{\eta,*}|| & \leq ||\Ub^\eta - \Ub^{L}||_{l^2(\mathbb{R}^M)} + ||\Ub^L - \Ub^{\eta,*}||_{l^2(\mathbb{R}^M)}  \\
    & \leq ||S\hat{\hb}_1^{L} - \Ub^{L}||_{l^2(\mathbb{R}^M)} + \frac{\varepsilon}{2} \quad (\textit{by Assumption \ref{asp:2}}) \\
    & \leq ||S\hat{\hb}_1^{L-1} - \Ub^{L-1}||_{l^2(\mathbb{R}^M)} + ||\hat{\mathcal{R}}(S\hat{\hb}_1^{L-1},\tilde{\Gb}) - \mathcal{R}(\Ub^{L-1},\tilde{\Gb})||_{l^2(\mathbb{R}^M)}  + \frac{\varepsilon}{2} \\
    & \leq ||S\hat{\hb}_1^{L-1} - \Ub^{L-1}||_{l^2(\mathbb{R}^M)} + 
     ||\hat{\mathcal{R}}(S\hat{\hb}_1^{L-1},\tilde{Gb}) - \mathcal{R}(S\hat{\hb}_1^{L-1},\tilde{Gb})||_{l^2(\mathbb{R}^M)} \\
     & + ||\mathcal{R}(S\hat{\hb}_1^{L-1},\tilde{Gb}) - \mathcal{R}(\Ub^{L-1},\tilde{Gb})||_{l^2(\mathbb{R}^M)} + \frac{\varepsilon}{2} \\
     & \leq (1+m)||S\hat{\hb}_1^{L-1} - \Ub^{L-1}||_{l^2(\mathbb{R}^M)} +  \frac{m\varepsilon}{2(1+m)^L}  + \frac{\varepsilon}{2}\quad \textit{(by Lemma \ref{lemma:1} and Assumption \ref{asp:1}})\\
     & \leq \frac{m\varepsilon}{2(1+m)^L}(1+(1+m)+(1+m)^2+\dots+(1+m)^{L-1}) + \frac{\varepsilon}{2} \\
     & \leq \frac{\varepsilon}{2} + \frac{\varepsilon}{2} = \varepsilon.
\end{align*}}

\end{proof}

\section{Formulation of Baseline Methods}\label{sec:gbml}

In this section, we discuss each baseline methods in details and how they are used in our experiments. A  meta-learning baseline in our problem setting would be to apply MAML and ANIL to a neural operator architecture. 
Here we formally state the implementation of ANIL and MAML for the problem described above, and they will serve as the baselinebaseline meta-based methods in our empirical experiments.

\textbf{MAML. }The MAML algorithm proposed in \citep{finn2017model} aims to find an initialization, $\tilde{\theta}$, across all tasks, so that new tasks can be learnt with very few gradient updates and examples. First, a batch $\{\mcT^\eta\}_{\eta=1}^H$ of $H$ tasks are drawn from the training task set. For each task $\mcT^\eta$, the context set of loading field/response field data pairs $\mcD^\eta$ is split to a support set of samples, $\mcS^\eta$, which will be used for inner loop updates, and a target set of samples, $\mcZ^\eta$, for outer loop updates. Then, for the inner loop, let $\theta^{\eta,0}:=\tilde{\theta}$ and $\theta^{\eta,i}$ be the task-wise parameter after $i$-th gradient update. During each inner loop update, the task-wise parameter is updated via
\begin{equation}\label{eqn:maml_inner}
\theta^{\eta,i}=\theta^{\eta,i-1}-\alpha \nabla_{\theta^{\eta,i-1}} \mcL_{\mcS^\eta}(\theta^{\eta,i-1}),\text{ for }\eta=1,\cdots,H,
\end{equation}
where $\mcL_{\mcS^\eta}(\theta^{\eta,i-1})$ is the loss on the support set of the $\eta$-th task, and $\alpha$ is the step size. After $m$ inner loop updates, the initial parameter $\tilde{\theta}$ is updated with a fixed step size $\beta$:
\begin{equation}\label{eqn:maml_outer}
\tilde{\theta}\leftarrow \tilde{\theta}-\beta \nabla_{\tilde{\theta}} \mcL_{\text{meta}}(\tilde{\theta}),
\text{ where the meta-loss }
\mcL_{\text{meta}}(\tilde{\theta}):=\sum_{\eta=1}^H\mcL_{\mcZ^\eta}(\theta^{\eta,m}).
\end{equation}
Then, on the test task, $\mcT^{\text{test}}$, an inner loop adaptation is performed based on few labelled samples $\mcD^{\text{test}}$ until convergence, and the approximated solution operator model is obtained on the test task as $\tilde{\mcG}[\gb;\theta^{\text{test}}]$.

\textbf{ANIL. }In \citep{raghu2019rapid}, ANIL was proposed as a modified version of MAML with inner loop updates only for the final layer. The inner loop update formulation of \eqref{eqn:maml_inner} is modified as
\begin{equation}\label{eqn:anil_inner}
\theta_Q^{\eta,i}=\theta_Q^{\eta,i-1}-\alpha \nabla_{\theta_Q^{\eta,i-1}} \mcL_{\mcS^\eta}(\theta_Q^{\eta,i-1}),\text{ for }\eta=1,\cdots,H,
\end{equation}
where $\theta_Q^{\eta,i}$ is the task-wise parameter on the final (projection) layer after $i$th gradient update. Then, the same outer loop updates are performed following \eqref{eqn:maml_outer}.

\textbf{Single/Pretrain1/Pretrain2. } We also implemented 3 non-meta-learning baseline approaches.
\begin{itemize}
\item \textbf{Single}: Learn a neural operator model only based on the context data set of the test task.
\item \textbf{Pretrain1}: Pretrain a  neural operator model based on all training task data sets, then fine-tune it based on the context test task data set.
\item \textbf{Pretrain2}: Pretrain a single neural operator model based on the context data set of one training task, then fine-tune it based on the context test task data set. To remove the possible dependency on the pre-training task, in this baseline we randomly select five training tasks for the purpose of pretraining and report the averaged results.
\end{itemize}
\section{Additional Results on Ablation Study}
\label{sec:ablation}

\begin{figure*}
     \centering
\includegraphics[width=0.7\textwidth]{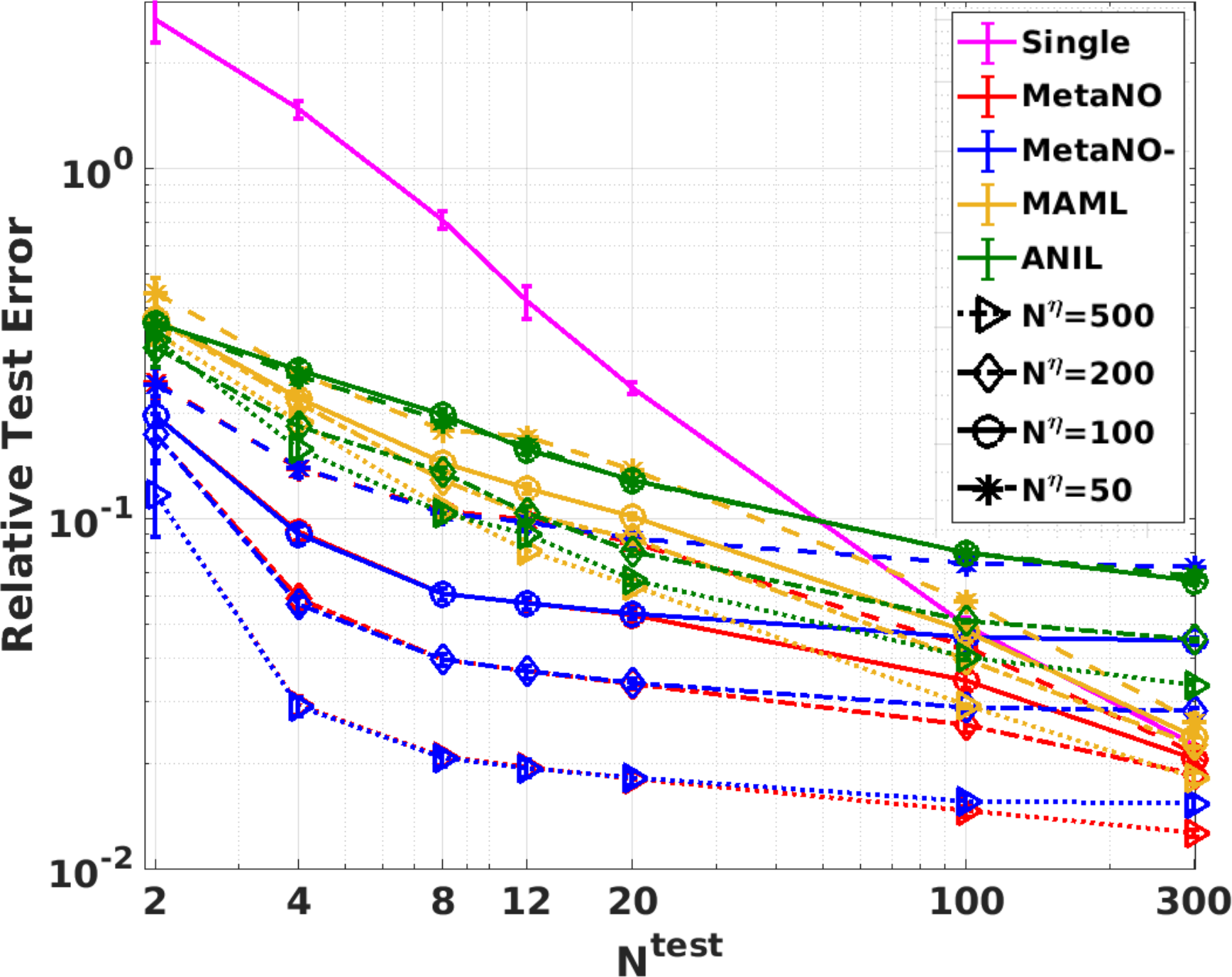}
\quad\includegraphics[width=0.75\textwidth]{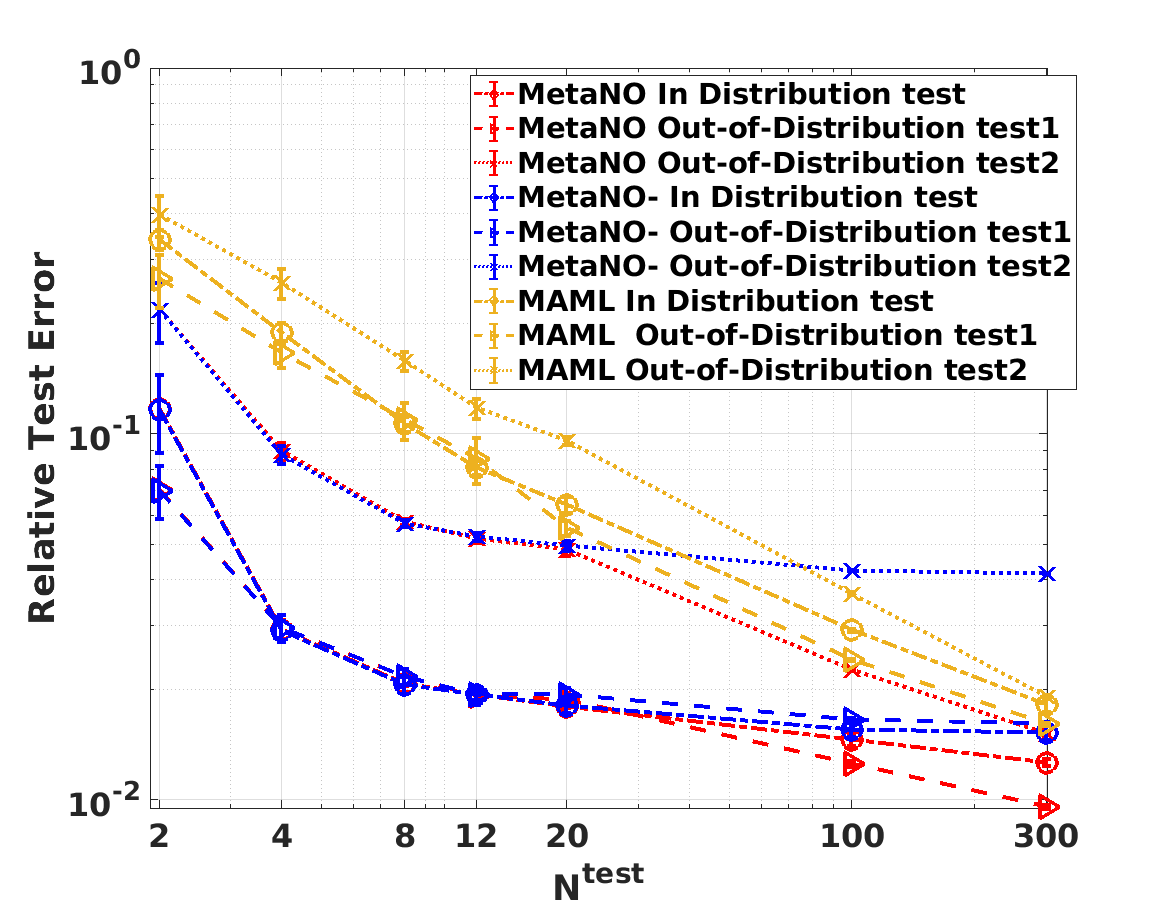}
     \caption{Additional results on a synthetic data set. Top: The full results showing the effect of varying training task context set sizes $N^\eta\in\{50,100,200,500\}$. Bottom: The relative error of MetaNO and MAML in in-distribution and out-of distribution tests.}
     \label{fig:metap_in_out}
 \end{figure*}
 
\textbf{Effect of Varying Training Context Set Sizes}
In this study, we  investigate the effect of different training task context sizes $N^\eta=\{50,100,200,500\}$ on four meta-learnt models: MetaNO, MetaNO-, MAML, and ANIL. The results are shown in Figure \ref{fig:metap_in_out}(Top). Here, MetaNO- and MetaNO did not have any inner loop updates. All parameters from all training tasks are optimized together. In MAML and ANIL we use half of the context set for inner loop updates (support set) and the other half for outer loop updates (target set). With the training task context size varying from $50$ to $500$, one can see that with more context data shown, all methods have improved performance, with decreasing relative test errors (with the same colors for the same methods across different context dataset). In addition, as the context set size in the test task grows, fine-tuning will gradually have better performance as MetaNO and MAML beats MetaNO- and ANIL, respectively. Overall MetaNO still achieve the best results.

\textbf{In-Distribution and Out-Of-Distribution Tests.} On bottom Figure \ref{fig:metap_in_out}, we demonstrate the relative test error of MetaNO against MAML in both ID and OOD tasks. 
We can see that test errors of these 3 tasks are  in a similar scale as the error on training tasks. The error from OOD task1 is comparable to the averaged ID test task error, while the error from OOD task2 is much larger, 
probably due to the fact that the solutions in OOD task1 generally have smaller magnitude and hence its solution operator lies more in a linear regime, which makes the solution operator learning task easier.
In all three cases, MetaNO outperforms MAML, 
hence validating the good generalization performance of MetaNO. Further details on the distribution of ID and OOD tasks as well as more discussions will be provided in Section \ref{app:eg1.1}.

\section{Data Generation and Training Details}
\label{sec:expdetails}
{ In the following we briefly describe the empirical process of generating datasets, and the settings employed in running of each algorithm. For a fair comparison, for each algorithm, we tune the hyperparameters, including the learning rate from $\{0.1,0.01,0.001,0.0001,0.00001,0.000001\}$, the decay rate from $\{ 0.5,0.7,0.9\}$, the weight decay parameter from $\{0.01,0.001,0.0001,0.00001,0.000001\}$, and the inner loop learning rate for MAML and ANIL from $\{0.01,0.001,0.0001,0.00001,0.000001\}$, to minimize the error on a separate validation dataset. In all experiments we decrease the learning rate with a ratio of learning rate decay rate every 100 epochs. The code and the processed datasets will be publicly released at Github for readers to reproduce the experimental results.}

\subsection{ Example 1: Synthetic Data Sets}

\subsubsection{ Data Generation}\label{app:eg1.1}


\begin{figure*}
    \centering
    \includegraphics[width=0.99\columnwidth]{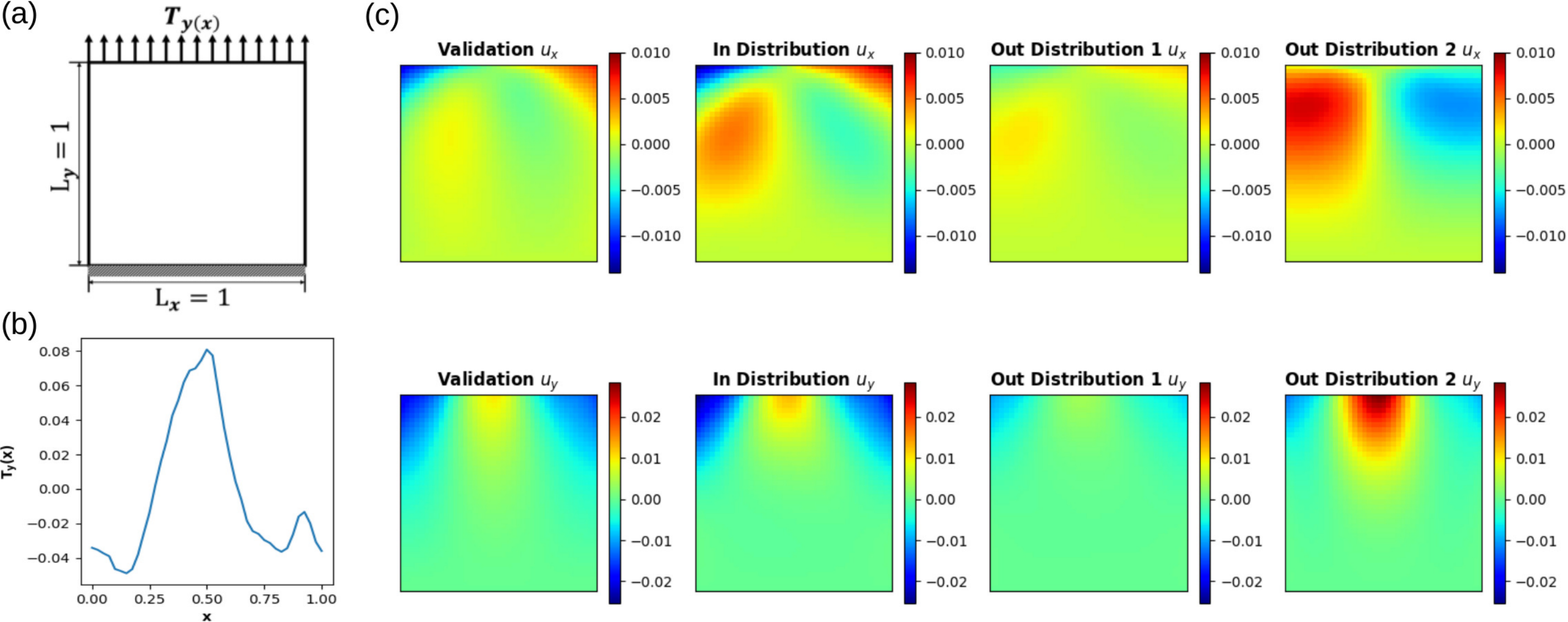}
    \caption{{ Problem setup of example 1: the synthetic data sets. (a) A unit square specimen subject to uniaxial tension with Neumann-type boundary condition. (b) \& (c) Visualization of an instances of the loading field $T_y(x)$, and the corresponding ground-truth solutions $\ub^\eta(\xb)$ from the in-distribution and out-of-distribution tasks, showing the solution diversity across different tasks, due to the change of underlying hidden material parameter set.}}
    \label{fig:hgoset}
\end{figure*}

{ In the synthetic data example, we consider the modeling problem of a hyperelastic, anisotropic, fiber-reinforced material, and seek to find its displacement field $\ub:[0,1]^2\rightarrow\real^2$ under different boundary loadings. In this problem,  the specimen is assumed to be subject to a uniaxial tension $T_y(\xb)$ on the top edge (see Figure \ref{fig:hgoset}(a)). To generate training and test samples, the Holzapfel-Gasser-Odgen (HGO) model~\citep{holzapfel2000new} was employed to describe the constitutive behavior of the material in this example, with its strain energy density function given as:
\begin{align*}
    \eta & = \frac{E}{4(1+\nu)}(\overline{I}_{1} - 2) - \frac{E}{2(1+\nu)}\ln(J)\\ 
    & +\frac{k_{1}}{2k_{2}}\left(\exp{(k_{2}\langle S(\alpha) \rangle^{2}}) + \exp{(k_{2}\langle S(-\alpha) \rangle^{2}}) - 2\right) + 
    \frac{E}{6(1-2\nu)}\left( \frac{J^{2} - 1}{2} - \ln{J} \right).
\end{align*}
Here, $\langle \cdot \rangle$ denotes the Macaulay bracket, and the fiber strain of the two fiber groups is defined as:
\begin{equation*}\label{eqn:fiberstrain}
    S(\alpha) = \frac{\overline{I}_{4}(\alpha) - 1 + |\overline{I}_{4}(\alpha) - 1|}{2}.
\end{equation*}
where $k_{1}$ and $k_{2}$ are fiber modulus and the exponential coefficient, respectively, $E$ is the Young's modulus for the non-fibrous ground matrix, and $\nu$ is the Poisson ratio. Moreover, $\overline{I}_{1}=\text{tr}(\mathbf{C})$ is the is the first invariant of the right Cauchy-Green tensor $\mathbf{C}=\mathbf{F}^T\mathbf{F}$, $\mathbf{F}$ is the deformation gradient, and $J$ is related with $\mathbf{F}$ such that $J = \det \mathbf{F}$. For the fiber group with angle direction $\alpha$ from the reference direction, $\overline{I}_{4}(\alpha)=\mathbf{n}^T(\alpha)\mathbf{C}\mathbf{n}(\alpha)$ is the fourth invariant of the right Cauchy-Green tensor $\mathbf{C}$, where $\mathbf{n}(\alpha)=[\cos(\alpha), \sin(\alpha)]^{T}$. To generate samples for different specimens,different specimens (tasks) correspond to different material parameter sets, $\{k_1,k_2,E,\nu,\alpha \}$. For the training tasks, the validation task, and the in-distribution (ID) test task, their physical parameters are sampled from: $k_1, k_2 \sim \mathcal{U}[0.1,1]$, $E \sim \mathcal{U}[0.55,1.5]$, $\nu \sim \mathcal{U}[0.01,0.49]$, and $\alpha \sim \mathcal{U}[\pi/10,\pi/2]$. For the two out-of-distribution (OOD) test tasks, we sample their parameters following $k_1, k_2 \sim \mathcal{U}[1,1.9]$, $E \sim 
\mathcal{U}[1.5,2] \cup \mathcal{U}[0.5,0.55]$, $\nu \sim \mathcal{U}[0.01,0.49]$\footnote{Here we sample both ID and OOD tasks from the same range of $\nu$, due to the fact that $[0.01,0.49]$ is the range of Poisson ratio for common materials \citep{bischofs2005effect}.}, and $\alpha \sim 
\mathcal{U}[\pi/2,3\pi/4]\cup[0,\pi/10]$. To generate the high-fidelity (ground-truth) dataset, we sampled $500$ different vertical traction conditions $T_y(\xb)$ on the top edge from a random field, following the algorithm in \cite{LangPotthoff2011,yin2022interfacing}. In particular, $T_y(\xb)$ is taken as the restriction of a 2D random field, $\phi(\xb) = \mathcal{F}^{-1}(\gamma^{1/2}\mathcal{F}(\Gamma))(\xb)$, on the top edge. Here, $\Gamma(\xb)$ is a Gaussian white noise random field on $\real^2$, $\gamma=(w_1^2+w^2_2)^{-\frac{5}{4}}$ represents a correlation function, and $w_1$, $w_2$ are the wave numbers on $x$ and $y$ directions, respectively. Then, for each sampled traction loading, we solved the displacement field on the entire domain by minimizing potential energy using the finite element method implemented in FEniCS \citep{alnaes2015fenics}. In particular, the displacement filed was approximated by continuous piecewise linear finite elements with triangular mesh, and the grid size was taken as $0.025$. Then, the finite element solution was interpolated onto $\chi$, a structured $41 \times 41$ grid which will be employed as the discretization in our neural operators.}

\begin{figure*}
    \centering
    \includegraphics[width=.49\textwidth]{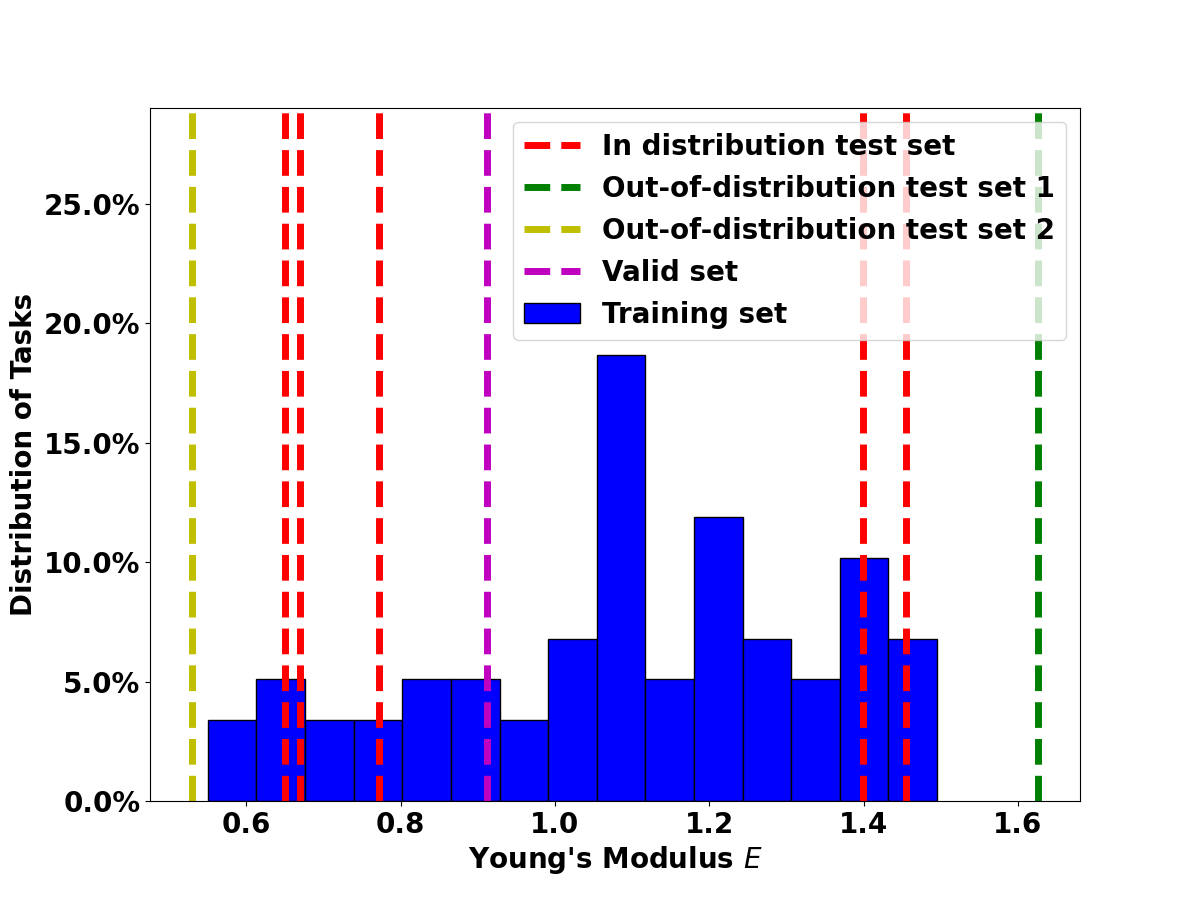}
    \includegraphics[width=.49\textwidth]{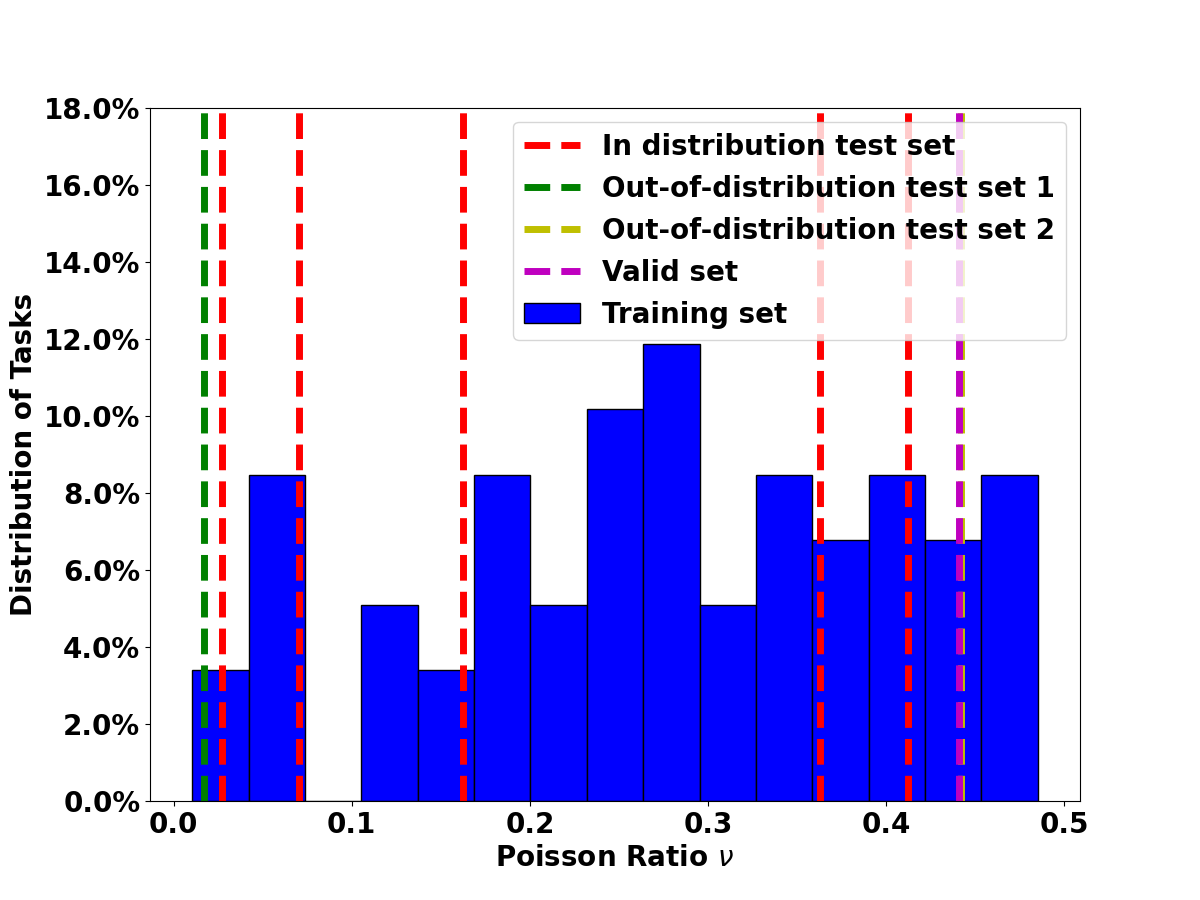}
     \includegraphics[width=.49\textwidth]{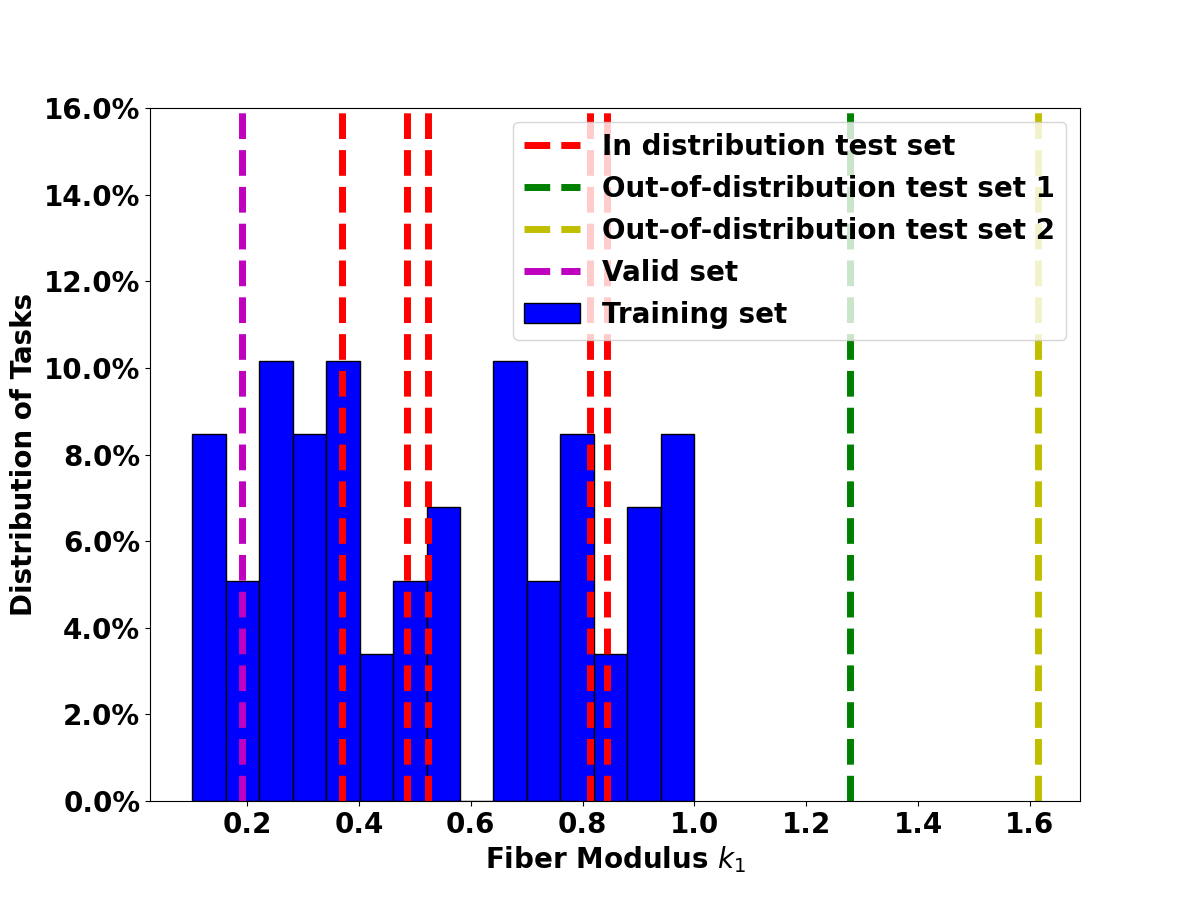}
 \includegraphics[width=.49\textwidth]{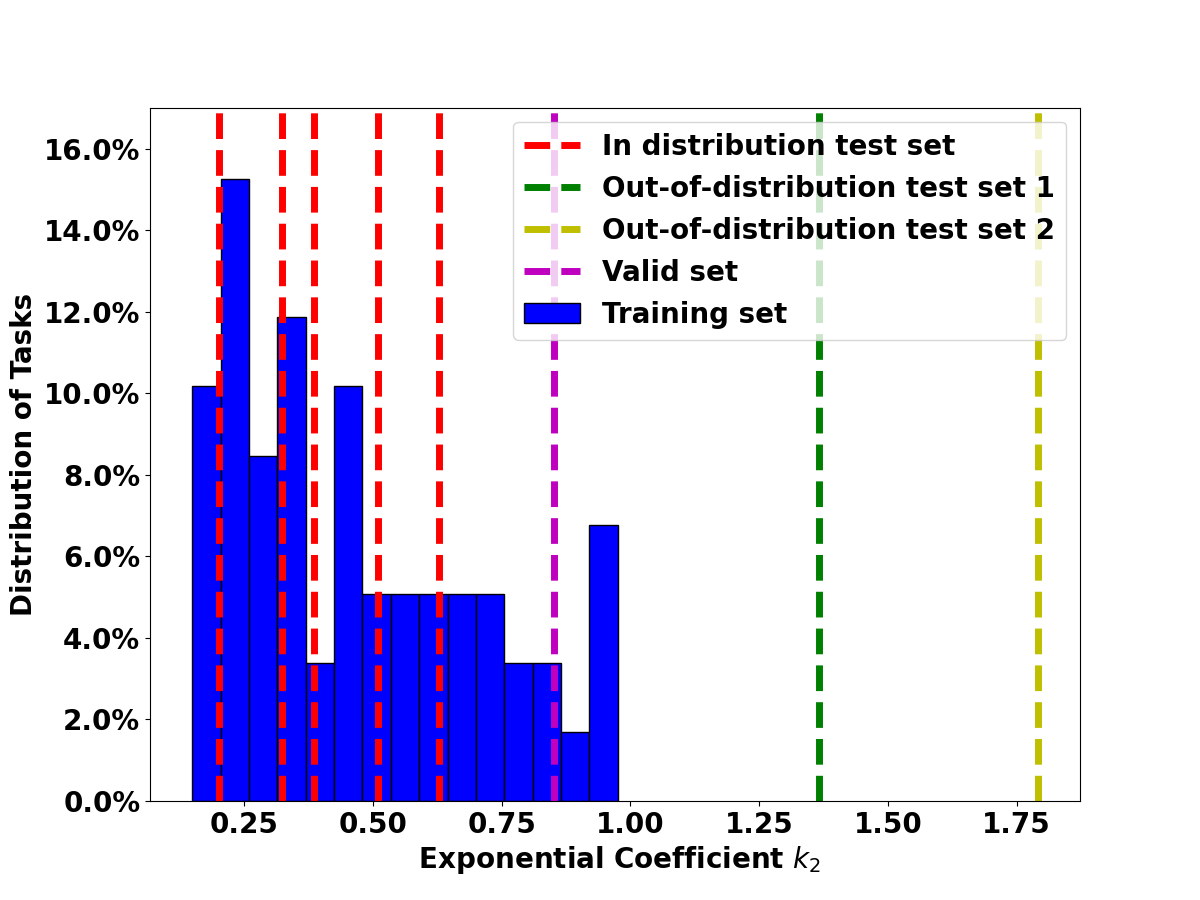}
    \includegraphics[width=.49\textwidth]{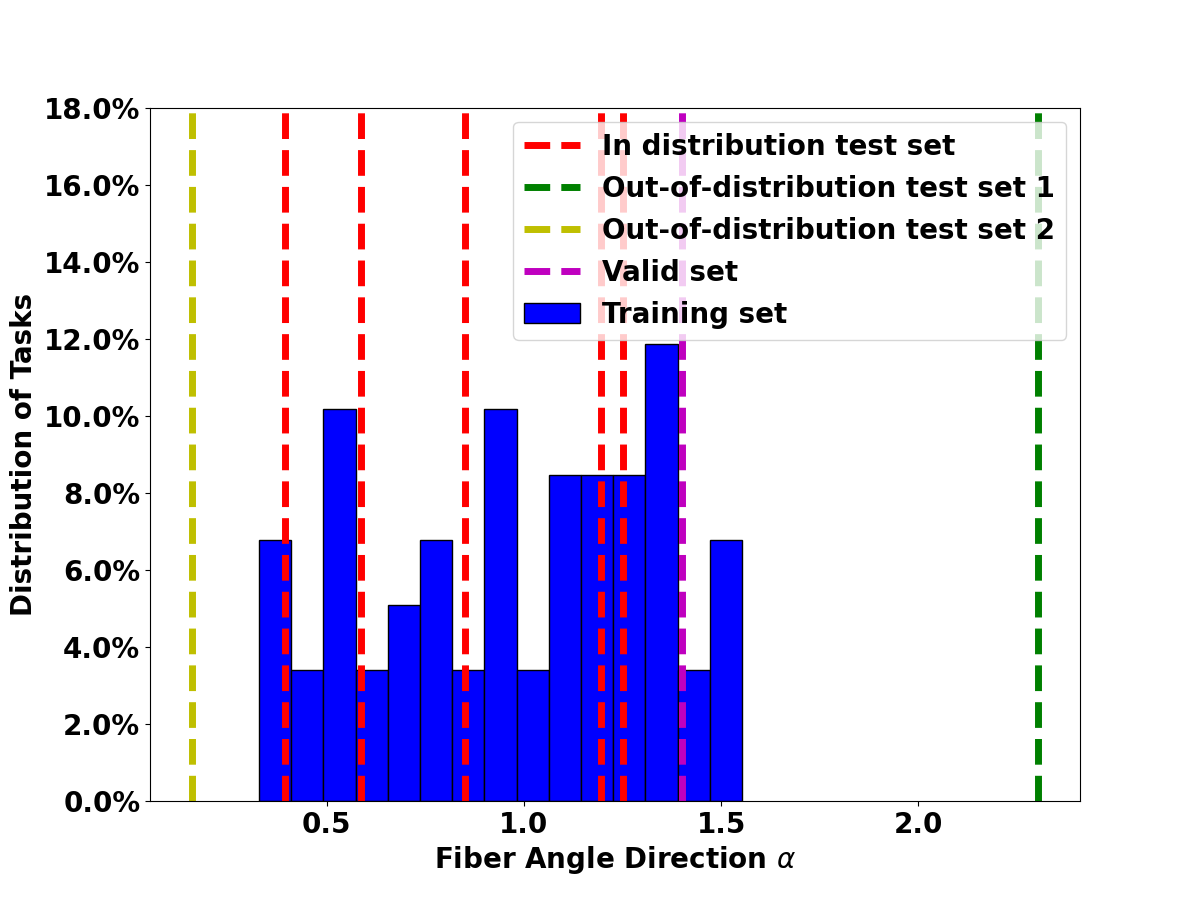}
 \includegraphics[width=.49\textwidth]{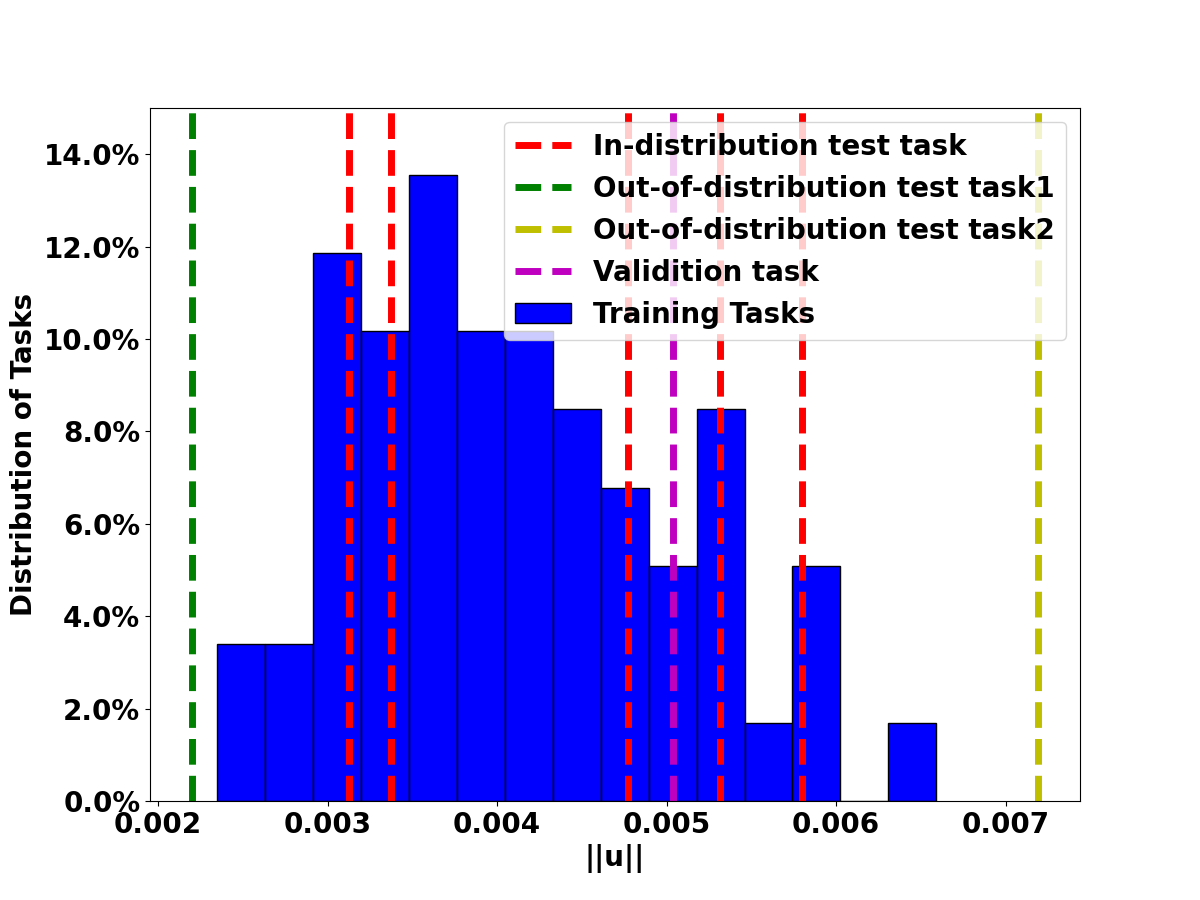}
    \caption{ Distribution of physical parameters of different tasks, and the resultant magnitude of material response, $\vertii{u^\eta(\xb)}_{L^2(\Omega)}$, on an exemplar loading instance shown in Figure \ref{fig:hgoset}(b).}
    \label{fig:distri}
\end{figure*}

{ To visualize the domain characteristics for tasks, the distribution of each parameter for training, validation and test tasks are demonstrated in Figure \ref{fig:distri}, and the corresponding solution fields are plotted in Figure \ref{fig:hgoset}(c), showing the diversity across different tasks due to the change of underlying hidden material parameter set, $\{k_1,k_2,E,\nu,\alpha \}$. From Figures \ref{fig:distri} and  \ref{fig:hgoset}(c), one can see that OOD Task1 corresponds to a stiffer material (with large Young's modulus $E$) and hence smaller deformation subject to the same loading $T_y(\xb)$. On the other hand, OOD Task2 corresponds to a softer material (with small Young's modulus $E$) and larger deformation. Therefore, the material response of OOD Task1 specimen is more likely to lie in a linear region, which is easier to learn and explains the relatively small test error on this task. On the other hand, the material response of OOD Task2 is more nonlinear and hence complex due to larger deformation, as shown in Figure \ref{fig:hgoset}(c), and results in the relatively larger test error in bottom Figure \ref{fig:metap_in_out}.}

\subsubsection{ Algorithm Hyperparameter Settings}


{ \textbf{Base model}: As the base model for all algorithms, we construct an  architecture for IFNO \citep{you2022learning} as follows. First, the input loading field instance $\gb(\xb)\in\mcA$ is lifted to a higher dimensional representation via lift layer $\mathcal{P}[\gb](\xb)$, which is parameterized as a 1-layer feed forward linear layer with width (3,32). Then for the iterative layer in \eqref{eq:IFNO}, we implement $\mathcal{F}^{-1}[\mathcal{F}[\kappa(\cdot;\vb)]\cdot \mathcal{F}[\hb(\cdot,l)]](\xb)$ with 2D fast Fourier transform (FFT) with input channel and output channel widths both set as 32 and the truncated Fourier modes set as 8. The local linear transformation parameter, $W$, is parameterized as a  1-layer feed forward network with width (32,32). In the projection layer, a 2-layer feed forward network with width (32,128,2) is employed. To accelerate the training procedure, we apply the shallow-to-deep training technique to initialize the optimization problem. In particular, we start from the NN model with depth $L=1$, train until the loss function reaches a plateau, then use the resultant parameters to initialize the parameters for the next depth, with $L=2$, $L=4$, and $L=8$. In the synthetic experiments, we set  the layer depth as $L=8$.}

{ \textbf{MetaNO}: 
During the meta-train phase, we train for the task-wise parameters $\theta_P^\eta$ and the common parameters $\theta_I$ and $\theta_Q$ on all 59 training tasks, with the context set of 500 samples on each task. After meta-train phase, we load $\theta_I$ and $\theta_Q$ and the averaged $\theta_P^\eta$ among all 59 tasks as initialization, then tune the hyperparameters based on the validation task. In particular, the 500 samples on the validation task is split into two parts: 300 samples are reserved for the purpose of training (as the context set) and the rest 200 samples are used for evaluation (as the target set). Then we train for the lift layer on the validation task, and tune the learning rate, the decay rate, and the weight decay parameter for different context set sizes ($N^{\text{test}}$), to minimize the loss on the target set. Based on the chosen hyperparameters, we perform the test on the test task by training for the lift layer on different numbers of samples on its context set, then evaluate and report the performance based on its target set. We repeat the procedure on the test task with selected hyperparameters with different 5 random seeds, and calculate  means and standard errors for the resultant test errors on target set.}

{ \textbf{MAML\&ANIL}: For MAML and ANIL, we use the same architecture as the base model, and also split the training tasks for the purpose of training (59 tasks) and validation (1 task) as in MetaNO. During the meta-train phase, for each task we randomly split the available 500 samples to two sets: 250 samples in the support set used for inner loop updates, and the rest in the target set for outer loop updates. During the inner loop update, we train for the task-wise parameter with one epoch, following the standard settings of MAML and ANIL~\citep{finn2017model,raghu2019rapid}. 
Then, the model hyperparameters, including the learning rate, weight decay, decay rate, and inner loop learning rate, are tuned. In the meta-test phase, we load the initial parameter and train for all parameters (in MAML) or the last-layer parameters (in ANIL) until the optimization algorithm converges. Similar as in MetaNO, we first tune the hyperparameters on the validation task, then evaluate the performance on the test task.} 


\subsection{ Example 2: Mechanical MNIST}

\subsubsection{ Data Settings}

{ Mechanical MNIST is a benchmark dataset of heterogeneous material undergoing large deformation, modeled by the Neo-Hookean material with a varying modulus converted from the MNIST bitmap images~\citep{lejeune2020mechanical}. In this example, we randomly select 1 specimen corresponding to each set of the hand-written numbers ``0'', ``2'', $\cdots$, ``9'', respectively, to obtain a set of 9 training tasks. Then, 6 randomly selected specimens from the set of  number ``1'' are used for validation (1 specimen) and test (5 specimens). On each specimen, we have 32 loading/response data pairs on a structured 27 by 27 grid, under the uniaxial extension, shear, equibiaxial extension, and confined compression load scenarios, respectively. 
On the validation and test tasks, we reserve a target set consisting of 20 data pairs for the purpose of evaluation, then use the rest as the context set.}

\subsubsection{ Algorithm Settings}
{ \textbf{Base model}: As the base model for all algorithms, we construct two IFNO architectures, for the prediction of $u_x$ and $u_y$, the displacement fields in the $x$- and $y$-directions, respectively. On each architecture, the input loading field instance $\gb(\xb)\in\mcA$ is mapped to a higher dimensional representation via a lifting layer $\mathcal{P}[\gb](\xb)$ parameterized as a 1-layer feed forward linear layer with width (4,64). Then for the iterative layer in \eqref{eq:IFNO}, we set the number of truncated Fourier mode as 13, and parameterize the local linear transformation parameter, $W$, as a 1-layer feed forward network with width (64,64). In the projection layer, a 2-layer feed forward network with width (64,128,1) is employed. In this example we also apply the shallow-to-deep technique to accelerate the training, and set the layer depth as $L=32$.}

{ \textbf{MetaNO}:  
During the meta-train phase, we train for the task-wise parameters $\theta_P^\eta$ and the common parameters $\theta_I$ and $\theta_Q$ on all 9 training tasks , with the context set of 32 samples on each task. After the meta-train phase, we load $\theta_I$ and $\theta_Q$ and the averaged $\theta_P^\eta$ among all 9 tasks as initialization, then train  $\theta_P$ on the validation task. In particular, the 32 samples on the validation task is split into two parts: 12 samples are reserved for the purpose of training (as the context set) and the rest 20 samples are used for the purpose of evaluation (as the target set). Then we train the lift layer on the validation task, and tune the learning rate, the decay rate, and the weight decay parameter for different  context set sizes ($N^{\text{test}}$), to minimize the loss on the target set. Based on the chosen hyperparameters, we perform the meta-test phase on the test task by training for the lift layer on different numbers of samples on its context set, then evaluate and report the performance based on its target set. We repeat the procedure with different 5 random seeds on each of the 5 test tasks, and calculate means and standard errors for the resultant test errors on the target set.}

{ \textbf{MAML\&ANIL}: For MAML and ANIL, we use the same architecture as the base model. 
During the meta-train phase, for each task we randomly split the available 32 samples to two sets: 16 samples in the support set used for inner loop updates, and the rest in the target set for outer loop updates. During the inner loop update, we also follow the standard settings of MAML and ANIL~\citep{finn2017model,raghu2019rapid}, and tune the hyperparameters following the same procedure as elaborated above for Example 1. }

\subsection{ Example 3: Experimental Measurements on Biological Tissues}

\subsubsection{ Data Generation}

\begin{figure*}
    \centering
    \includegraphics[width=0.9\columnwidth]{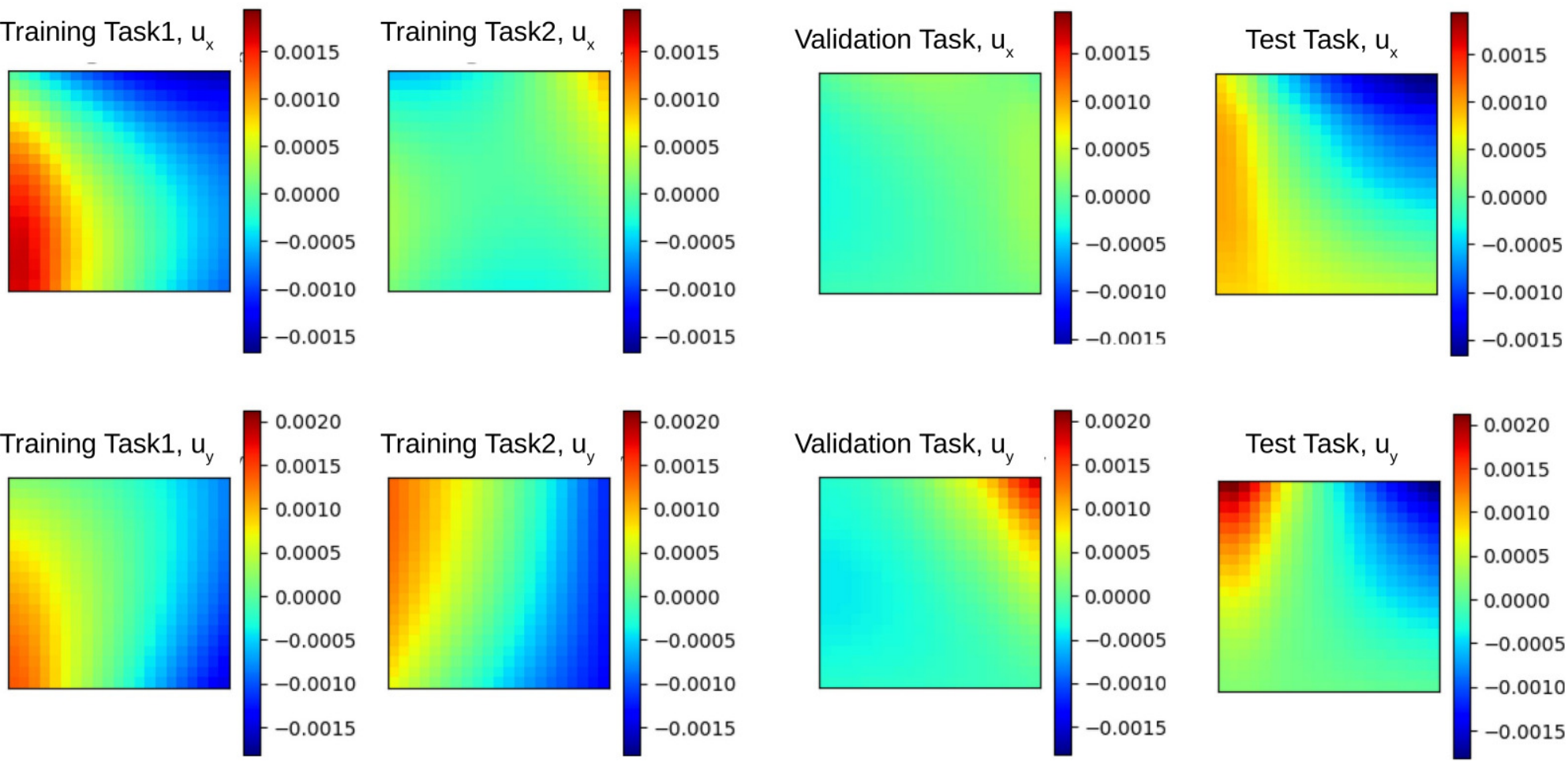}
    \caption{{ Visualization of the processed dataset in example 3: learning the biological tissue responses. Subject to the same loading instance, different columns show the corresponding ground-truth solutions $\ub^\eta(\xb)$ from different tasks, showing the solution diversity across different tasks due to the change of underlying hidden material parameter field.}}
    \label{fig:tissueplot}
\end{figure*}

{ We now briefly provide the data generation procedure for the tricuspid valve anterior leaflet (TVAL) response modeling example. In this problem, the constitutive equations and material microstructure are both unknown, and the dataset has unavoidable measurement noise. To generate the data, we firstly followed the established biaxial testing procedure, including acquisition of a healthy porcine heart and retrieval of the TVAL \cite{ross2019investigation,laurence2019investigation}. Then, we sectioned the leaflet tissue and applied a speckling pattern to the tissue surface using an airbrush and black paint \cite{zhang2004applications,lionello2014practical,palanca2016use}. The painted specimen was then mounted to a biaxial testing device (BioTester, CellScale, Waterloo, ON, Canada). To generate samples for each specimen, we performed 7 protocols of displacement-controlled testing to target various biaxial stresses: $P_{11}:P_{22}=\{1:1, 1:0.66, 1:0.33, 0.66:1, 0.33:1, 0.05:1, 1:0.1\}$. Here, $P_{11}$ and $P_{22}$ denote the first Piola-Kirchhoff stresses in the $x$- and $y$-directions, respectively. Each stress ratio was performed for three loading/unloading cycles. Throughout the test, images of the specimen were captured by a CCD camera, and the load cell readings and actuator displacements were recorded at 5\,Hz. After testing, the acquired images were analyzed using the digital image correlation (DIC) module of the BioTester's software. The pixel coordinate locations of the DIC-tracked grid were then exported and extrapolated to a 21 by 21 uniform grid.}

{ In this example, we have the DIC measurements on 16 specimens, with 500 data pairs of loadings and material responses from the 7 protocols on each specimen. These specimens are divided into three groups: 12 for the purpose of meta-train, 1 for validation, and 3 for test. To demonstrate the diversity of these specimens due to the material heterogeneity in biological tissues, in Figure \ref{fig:tissueplot} we plot the processed displacement field of two exemplar training specimens and the validation and test specimens. For each model, the results are reported as the average of all 3 test tasks. }

\subsubsection{ Algorithm Settings}

{ \textbf{Base model}: As the base model, we first construct the lifting layer as a 1-layer feed forward linear layer with width (4,16). Then for the iterative layer in we keep 8 truncated Fourier modes and parameterize the local linear transformation parameter, $W$, a 1-layer feed forward network with width (16,16). In the projection layer, a 2-layer feed forward network with width (16,64,1) is employed. We construct two 4-layer IFNO architectures, for the prediction of $u_x$ and $u_y$, the displacement fields in the $x$- and $y$-directions, respectively.}

{ \textbf{MetaNO}:  
During the meta-train phase, we train for the task-wise parameters $\theta_P^\eta$ and the common parameters $\theta_I$ and $\theta_Q$ on all 12 tasks, with the context set of 500 samples on each task. After meta-train phase, we load $\theta_I$ and $\theta_Q$ and the averaged $\theta_P^\eta$ among all 12 tasks as initialization, then tune the hyperparameters based on the validation task. In particular, the 500 samples on the validation task is divided into two parts: 300 samples are reserved for the purpose of training (as the context set) and the rest 200 samples are used for evaluation (as the target set). 
Based on the chosen hyperparameters, we perform the test on the test tasks by training for the lift layer on different numbers of samples on its context set, then evaluate the performance based on its target set. 
}

{ \textbf{MAML\&ANIL}: For MAML and ANIL, we use the same architecture as the base model, and also split the training tasks for the purpose of training and validation as in MetaNO. During the meta-train phase, for each task we randomly split the available 500 samples to two sets: 250 samples in the support set used for inner loop updates, and the rest in the target set for outer loop updates. During the inner loop update, we train for the task-wise parameter with one epoch, following the standard settings of MAML and ANIL~\citep{finn2017model,raghu2019rapid}. }



\end{document}